%% file: main.tex
\documentclass[letterpaper]{article}
\usepackage[preprint]{aaai2027}
\usepackage[hyphens]{url}
\usepackage{graphicx,natbib,caption}
\usepackage{booktabs,multirow,amsmath,amssymb}
\usepackage{microtype}
\title{Why Do VLMs Fail at Remote Physiological Measurement? Rethinking with Physiological Grounding and Numerical Alignment}
\author{
    Xiao Yang\textsuperscript{\rm 1},
    Zekun Zhang\textsuperscript{\rm 2},
    Yuxuan Fan\textsuperscript{\rm 3},
    Duoxun Tang\textsuperscript{\rm 4},\\
    Ziyang Liu\textsuperscript{\rm 1},
    Dengbo He\textsuperscript{\rm 1},
    Kaishun Wu\textsuperscript{\rm 1}\corresponding,
    Jiyao Wang\textsuperscript{\rm 1}
}

\affiliations{
    \textsuperscript{\rm 1}The Hong Kong University of Science and Technology (Guangzhou), China\\
    \textsuperscript{\rm 2}Sichuan University, China\\
    \textsuperscript{\rm 3}Nanyang Technological University, Singapore\\
    \textsuperscript{\rm 4}Tsinghua Shenzhen International Graduate School, Tsinghua University, China
}

\begin{document}

\maketitle
\pagestyle{plain}
\thispagestyle{plain}

\begin{abstract}
Video-based remote physiological measurement (RPM) has broad applications, while Vision-Language Models (VLMs) provide a promising opportunity to enable end-to-end physiological monitoring for future human-aware intelligent systems. However, existing RPM methods rely on task-specific models and cannot be seamlessly integrated with the multimodal understanding and interaction capabilities of VLMs. We revisit RPM with VLMs and provide empirical evidence for two major bottlenecks: generic visual semantic representations fail to preserve fine-grained physiological cues; while discrete token prediction in autoregressive generation is inherently misaligned with continuous numerical measurement. To address these challenges, we propose VITAL (Visual-physiological Token Alignment for Language Models), a visual-physiological alignment framework for HR estimation and general video understanding. VITAL retains the original visual pathway for semantic understanding and introduces an independent physiological pathway to jointly model semantic and physiological information. Moreover, we propose a structured numerical alignment approach to improve the reliability of generative heart rate estimation by enhancing numerical relationship modeling. Extensive experiments show that VITAL improves cross-domain estimation performance on multiple rPPG datasets while largely retaining the VLM's general video-understanding capabilities.
\end{abstract}

\begin{links}
    \link{Code}{https://github.com/XiaoYANG274/VITAL}
\end{links}

\begin{figure}[t]
    \centering
    \includegraphics[width=\columnwidth]{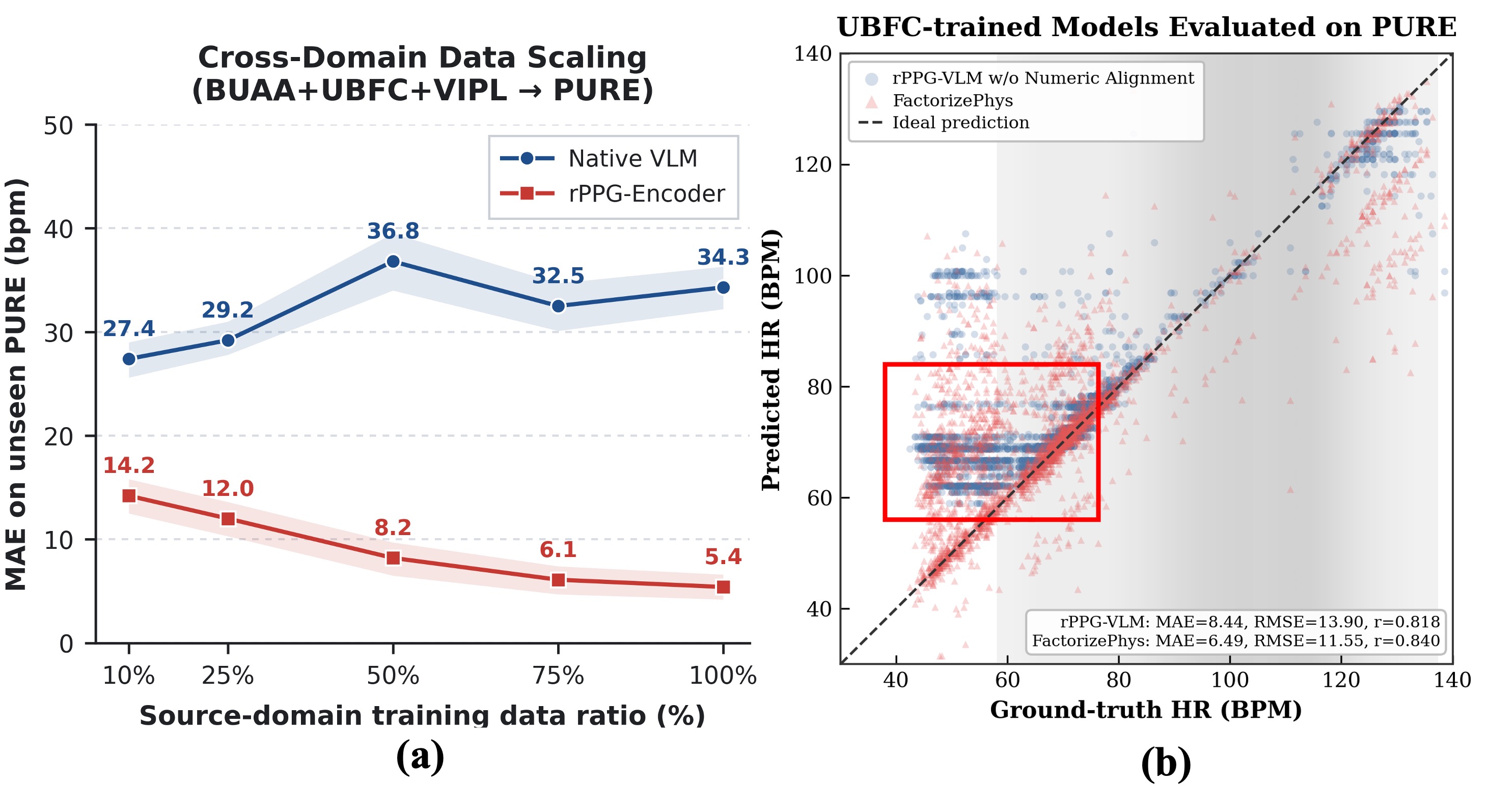}
    \caption{
    Empirical test of VLM-based RPM.
    (a) Increasing source-domain training data does not improve the cross-domain generalization of native VLMs, whereas an rPPG-oriented physiological encoder (FactorizePhys) yields consistent gains.
    (b) Even with physiological representations, generative VLMs exhibit source-prior attraction under HR distribution shift. Models are trained on UBFC-rPPG and evaluated on PURE; darker gray indicates higher source-domain HR density. Red boxes highlight repeated-value banding.
    }
    \label{fig:empirical_test}
\end{figure}

\section{Introduction}

Remote photoplethysmography (rPPG) enables contactless estimation of physiological measurements such as heart rate (HR) by analyzing subtle skin-color variations in facial RGB videos \cite{verkruysse2008remote}. Recent deep learning methods have substantially improved rPPG accuracy and robustness \cite{wang2026rpm,wang2024condiff}, supporting applications in driver monitoring \cite{wang2024efficient}, human--computer interaction \cite{wang2026mild}, and everyday health monitoring \cite{liao2026passive,wu2025ubicon,hu2026beyond}. Meanwhile, vision-language models (VLMs) have shown strong general-purpose capabilities in video understanding~\cite{lin2024video,tao2025moss}, multimodal reasoning, and natural language interaction \cite{chen2025blip3,fan2026oralgpt}. Equipping VLMs with remote physiological measurement (RPM) capabilities could therefore provide a unified interface for physiological sensing and human-aware interaction. Recent large-model-based approaches either still substantially underperform dedicated rPPG models \cite{zhang2025period}, or mainly employ VLMs for frequency-aware representation learning, prior transfer, or as components of task-specific physiological pipelines \cite{yue2025bootstrapping,gao2026physvlm,xie2025physllm,lu2024gptpsychologist}. This raises a fundamental question: Can a general-purpose VLM reliably perform end-to-end RPM while largely retaining its semantic understanding and language interaction capabilities?

To answer this question, we first examine whether increasing rPPG instruction data can overcome the limitation. Cross-domain performance fails to improve as the training set grows. Then, if we keep the language model, supervision, and training procedure unchanged, replacing the native visual encoder with an rPPG-oriented physiological encoder substantially improves HR estimation and restores consistent gains with increasing data scale. Given these results (Fig.~\ref{fig:empirical_test} (a)), we argue \emph{visual--physiological representation mismatch} is one of the primary causes: visual encoders pretrained for semantic understanding may fail to preserve the fine-grained spatiotemporal cues required for rPPG. However, replacing the native encoder leads to a decline in generic vision understanding capability, motivating the joint modeling of semantic and physiological evidence.

Even with a physiological encoder, generative rPPG-VLMs tend to contract their predictions toward high-density regions of the source-domain HR distribution under dataset shift (Fig.~\ref{fig:empirical_test} (b)). We refer to this phenomenon as \emph{source-prior attraction}. Standard token-wise and sample-wise cross-entropy treats numerical outputs as discrete token sequences without explicitly modeling numerical distance, decimal place value, or cross-sample ordering. The model may therefore reproduce frequent source-domain values rather than track the physiological variation in the input video.

With the above two challenges, we propose Visual-Physiological Token Alignment for Language Models (VITAL). VITAL preserves the native semantic visual pathway and adds a parallel physiological pathway that maps fine-grained spatiotemporal features into Physiological Tokens (PhysTokens). Structured Numerical Alignment further constrains HR generation through place-value-aware numerical supervision within individual samples and ordinal supervision across samples. All additional objectives are used only during training, while inference remains fully generative through the original language generation head. Our main contributions are as follows:

\begin{itemize}

\item Through controlled scaling and encoder-replacement experiments, we identify two key bottlenecks in applying general-purpose VLMs to rPPG: a visual--physiological representation mismatch and a discrete--continuous numerical interface mismatch. We further characterize source-prior attraction as a distinct failure pattern under cross-dataset HR distribution shift.

\item We propose VITAL, which combines Dual-Path Physiological Grounding with Structured Numerical Alignment. The former preserves semantic visual representations while introducing PhysTokens as fine-grained physiological evidence, and the latter aligns discrete digit generation with the continuous HR space through place-value-aware and cross-sample ordinal supervision.

\item Experiments on six rPPG datasets show that VITAL improves cross-dataset HR estimation and generalization to challenging conditions, while largely preserving the general video understanding capability of the base VLM.

\end{itemize}

\section{Related Work}

\subsection{Remote Physiological Measurement}

Remote photoplethysmography (rPPG) estimates physiological indicators such as HR from subtle skin-color variations in facial RGB videos. Early methods relied on handcrafted color transformations and signal projections \cite{verkruysse2008remote,de2013robust}, whereas recent approaches learn spatiotemporal physiological representations directly from videos \cite{wang2026rpm,lu2025rhythmguassian,yang2025not}. Despite improved accuracy and robustness, these methods remain task-specific and typically require dedicated waveform-recovery modules or regression heads~\cite{wang2025physmle,wang2026gap}, limiting their integration with general video understanding and language interaction.

\subsection{Large Models for Physiological Perception}

Recent studies have explored vision--language and language models for remote physiological perception~\cite{lu2026sagedeer}. VL-phys adapts a pretrained VLM with frequency-oriented vision--text pairs for self-supervised rPPG representation learning \cite{yue2025bootstrapping}, PhysVLM distills frequency-related VLM priors into a visual-only multitask student \cite{gao2026physvlm}, and PhysLLM uses a language model to enhance dedicated physiological representations \cite{xie2025physllm}. These methods demonstrate the value of language-supervised frequency priors and large-model representations, but primarily employ large models as representation learners, teachers, or components of task-specific pipelines. Period-LLM directly generates physiological estimates by introducing periodic visual modeling into a multimodal language model \cite{zhang2025period}; however, its reported performance remains behind dedicated RPM models, and the cross-domain behavior and failure mechanisms of generative RPM-VLMs remain insufficiently understood. We address this gap by examining and improving physiological representation and autoregressive numerical generation under domain shift while retaining the native semantic pathway and language-generation interface.

\section{Empirical Test: Why Do VLMs Fail at rPPG?}
\label{sec:diagnosis}

In this section, we conduct controlled experiments to investigate 1) whether data scaling can overcome this limitation; 2) whether the bottleneck
originates from visual representations; and 3) how generative prediction behaves under cross-dataset HR distribution shift.

\subsection{More rPPG Instructions Are Not Enough}
\label{sec:data_scaling}

We first adopt the most direct rPPG instruction-tuning setup. Supervised fine-tuning is performed using the standard autoregressive language modeling objective. To study scaling behavior, we construct nested training subsets
$\mathcal{D}_{\rho}$ from source-domain rPPG datasets, where
$\rho\in(0,1]$ denotes the fraction of training data used. Across all data scales, we use the same model initialization, video sampling strategy, instruction templates, and optimization settings, changing only the number of supervised training samples.

As shown by the blue curve in Fig.~\ref{fig:empirical_test} (a), increasing
rPPG instruction data does not improve the cross-domain performance of
the vanilla VLM, whose predictions remain weakly correlated with
ground-truth HR. This indicates that insufficient supervision
is not the only bottleneck.

\subsection{Localizing the Physiological Representation Bottleneck}
\label{sec:representation_bottleneck}

We replace the native visual encoder $E_{\mathrm{sem}}$ with FactorizePhys $E_{\mathrm{phys}}$, pretrained using only source-domain physiological supervision. No target-domain data are used for training or model selection.

To ensure a controlled comparison, both encoders use identical 30-fps video clips, frame counts, spatial resolutions, and facial preprocessing, while all language-side components and training settings remain unchanged. As shown by the red curve in Fig.~\ref{fig:empirical_test} (a), the physiological encoder substantially improves HR estimation and restores consistent gains with increasing data scale, localizing a major bottleneck to the encoder-side representation.

We refer to this problem as a \emph{visual--physiological representation mismatch}. General-purpose visual encoders are primarily pretrained to capture high-level semantics such as objects, actions, and scenes. This interpretation is consistent with recent evidence that contemporary VLMs are largely insensitive to frame-order shuffling and often rely on order-agnostic aggregation of frame-level appearance rather than temporally ordered visual evidence \cite{Agarwal_2026_ECCV}. Such temporal-order insensitivity is particularly problematic for remote HR estimation, which requires subtle cross-frame color variations and periodic spatiotemporal patterns in skin regions to be preserved. A visual representation that supports general video understanding is therefore not necessarily suitable for remote physiological measurement.

Although the replacement experiment cannot isolate the effect of a
single architectural factor, it shows that directly replacing the native
encoder improves rPPG at the cost of losing semantic visual information.
This motivates a dual-path design that retains both semantic and
physiological evidence.

\subsection{Source-Prior Attraction under Dataset Shift}
\label{sec:source_prior}

Although physiological representations improve basic HR prediction,
generative rPPG-VLMs still exhibit numerical generalization problems
under dataset shift. We train the physiological-encoder-based rPPG-VLM
on UBFC-rPPG \cite{bobbia2019unsupervised} and evaluate it on PURE
\cite{stricker2014non}, using FactorizePhys
\cite{joshi2024factorizephys} as a task-specific regression baseline.
Since the native visual encoder has already been replaced, this setting
focuses on numerical prediction rather than physiological representation
quality.

As shown in Fig.~\ref{fig:empirical_test}(b), the rPPG-VLM produces
reasonable estimates within the source-domain HR range. However, when
target HR values extend beyond the observed distribution, its predictions
contract toward high-density source-domain regions, whereas the
task-specific regression model preserves a broader dynamic range. We
refer to this phenomenon as \emph{source-prior attraction}. A more severe
case, in which predictions collapse into a narrow source-domain region,
is termed \emph{source-prior collapse}.

To isolate the effect of the numerical output interface, we compare two
decoders branching from the same projector-pretrained physiological
representation. One predicts HR through a linear regression head, whereas
the other generates HR autoregressively through the original language
head. Both are trained on UBFC-rPPG and evaluated on PURE.

\begin{table}[!ht]
\centering
\begingroup
\small
\setlength{\tabcolsep}{2pt}
\begin{tabular}{@{}lccccccc@{}}
\toprule
\multirow{2}{*}{Decoder}
& \multicolumn{3}{c}{Estimation}
& \multicolumn{4}{c}{Collision rate (\%)$\downarrow$} \\
\cmidrule(lr){2-4}
\cmidrule(lr){5-8}
& MAE$\downarrow$ & RMSE$\downarrow$ & $r\uparrow$
& $\Delta\mathrm{GT}\geq5$
& $\geq10$
& $\geq15$
& $\geq20$ \\
\midrule
Reg.
& 9.57 & 13.85 & 0.900
& 6.01 & 3.62 & 2.20 & 1.19 \\
AR
& 8.44 & 13.90 & 0.818
& 7.12 & 6.55 & 5.77 & 4.02 \\
\bottomrule
\end{tabular}
\endgroup
\caption{Output-interface analysis on
UBFC-rPPG$\rightarrow$PURE. Collision rates are the percentages of
sample pairs with $|\Delta\mathrm{Pred}|\leq1$ BPM among pairs satisfying
the indicated minimum ground-truth HR separation. Reg.: linear regression; AR: autoregressive language output.}
\label{tab:numerical_collision}
\end{table}

As shown in Tab.~\ref{tab:numerical_collision}, autoregressive
generation achieves slightly lower MAE but exhibits lower correlation
and consistently higher pairwise collision rates. The difference becomes
more pronounced for physiologically distinct samples: among pairs whose
ground-truth HR values differ by at least 20 BPM, 4.02\% receive
predictions within 1 BPM under autoregressive generation, compared with
1.19\% under continuous regression. This result suggests that standard
token-wise supervision weakens cross-sample numerical discriminability
despite competitive pointwise error, because it does not explicitly
encode numerical distance, place value, or ordinal relationships.

Together, these experiments reveal two interface mismatches: one between visual representations and physiological evidence, and the other between autoregressive number-token generation and continuous numerical measurement.

\begin{figure*}[t]
    \centering
    \includegraphics[width=\textwidth]{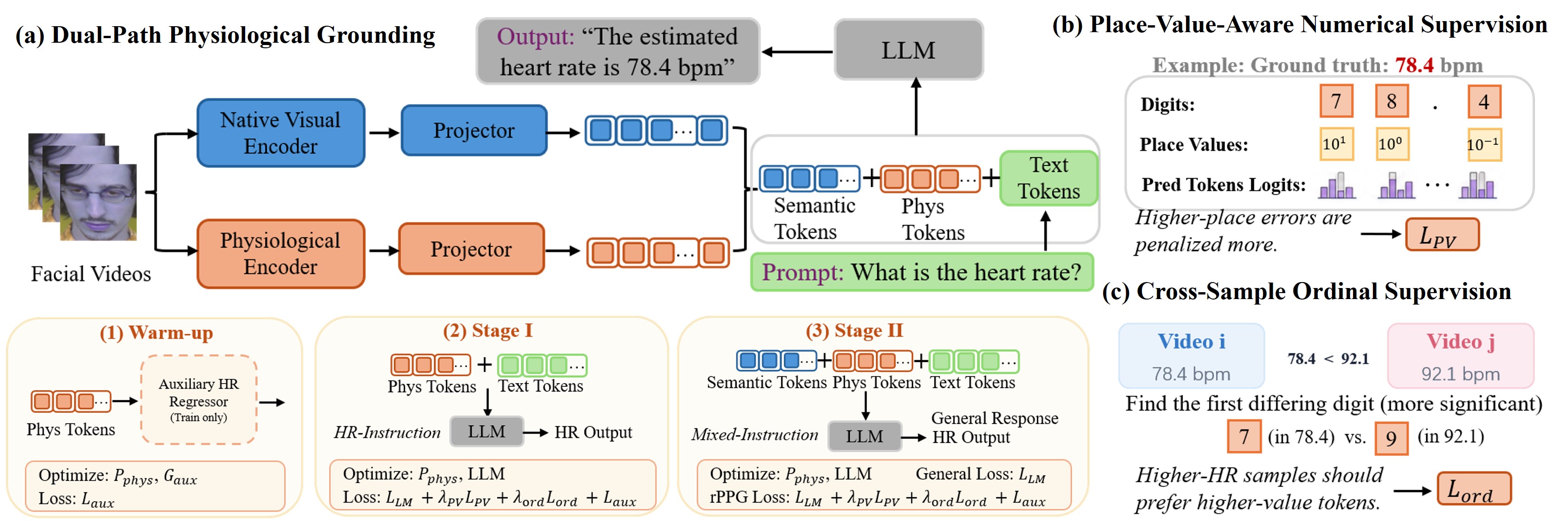}
    \caption{Overview of VITAL.
(a) Dual-Path Physiological Grounding preserves native semantic representations while introducing PhysTokens as fine-grained physiological evidence, together with a progressive alignment training process.
(b) Place-Value-Aware Numerical Supervision weights digit errors according to decimal significance.
(c) Cross-Sample Ordinal Supervision preserves relative HR ordering through the first differing digit.}
    \label{fig:vital_overview}
\end{figure*}

\section{Method}
\label{sec:method}

\subsection{Overview}
\label{sec:method_overview}

Given a facial video $X_i$ and a text instruction $q_i$, VITAL generates
a natural-language response $Y_i$ containing the HR estimate.
It consists of Dual-Path Physiological Grounding, which preserves the
native semantic pathway while introducing PhysTokens from an rPPG-oriented
encoder, and Structured Numerical Alignment, which constrains digit
generation through place-value-aware and cross-sample ordinal supervision.
An overview of VITAL is shown in Fig.~\ref{fig:vital_overview}.

\subsection{Dual-Path Physiological Grounding}
\label{sec:dual_path}

\paragraph{Semantic and physiological pathways.}
As shown in Fig.~\ref{fig:vital_overview}(a), VITAL retains the pretrained semantic encoder $E_{\mathrm{sem}}$ and projector $P_{\mathrm{sem}}$, and adds a physiological encoder $E_{\mathrm{phys}}$ with a lightweight projector $P_{\mathrm{phys}}$ to map physiological features into the language-model space:
\begin{equation}
\begin{aligned}
Z_i^{\mathrm{sem}}
&=P_{\mathrm{sem}}\!\left(E_{\mathrm{sem}}(X_i)\right)
\in\mathbb{R}^{N_s\times d},\\
Z_i^{\mathrm{phys}}
&=P_{\mathrm{phys}}\!
\!\left(E_{\mathrm{phys}}(X_i)\right)
\in\mathbb{R}^{N_p\times d}
\end{aligned}
\label{eq:dual_tokens}
\end{equation}
where $N_s$ and $N_p$ are the numbers of semantic and physiological tokens, and $d$ is the language-model hidden dimension. $Z_i^{\mathrm{sem}}$ provides semantic context about people, actions, scenes, occlusions, and lighting conditions, while $Z_i^{\mathrm{phys}}$ (PhysTokens) encodes skin-color variations, local physiological consistency, and quasi-periodic temporal patterns as physiological evidence rather than final HR values.

The native semantic visual tokens, PhysTokens, and text instruction jointly form the conditioning sequence of the language model:
\begin{equation}
U_i=
\left[
Z_i^{\mathrm{sem}};
Z_i^{\mathrm{phys}};
E_{\mathrm{text}}(q_i)
\right]
\label{eq:dual_fusion}
\end{equation}

$E_{\mathrm{text}}$ is the text embedding layer. Conditioned on $U_i$, the language model generates the response autoregressively.

\paragraph{Progressive PhysToken Alignment.}
Although the pretrained physiological encoder provides effective physiological representations, its features are not yet aligned with the language-model space. After warming up the physiological projector with auxiliary regression, we align PhysTokens with the language-model space in two stages. Stage I uses only PhysTokens and text, $U_i^{(1)}=[Z_i^{\mathrm{phys}};E_{\mathrm{text}}(q_i)]$, to learn textual HR prediction. Stage II initializes from Stage I and introduces native visual tokens through the full input $U_i$ for joint adaptation. This progression prevents premature reliance on the mature semantic pathway while ignoring unaligned PhysTokens. Stage II also mixes general-purpose video instructions at a fixed ratio to mitigate catastrophic forgetting.

To preserve decodable continuous HR information during this mapping process, we attach an auxiliary HR regression head to the PhysTokens for training only:
\begin{equation}
\begin{aligned}
\widehat{h}_i^{\mathrm{aux}}
&=
G_{\mathrm{aux}}\!\left(
\operatorname{Pool}\!\left(Z_i^{\mathrm{phys}}\right)
\right),\\
\mathcal{L}_{\mathrm{aux}}
&=
\frac{1}{B}
\sum_{i=1}^{B}
\left(
\widehat{h}_i^{\mathrm{aux}}-h_i
\right)^2
\end{aligned}
\label{eq:aux_loss}
\end{equation}
Here, $B$ is the number of HR-annotated physiological samples in the mini-batch. During warm-up, the physiological encoder is frozen while the projector and auxiliary head are jointly optimized. During VLM adaptation, the encoder and auxiliary head remain frozen; $\mathcal{L}_{\mathrm{aux}}$ updates the projector through the fixed head to preserve decodable continuous HR information. At inference, the auxiliary head is removed and HR is generated through the language head.

\subsection{Structured Numerical Alignment}
\label{sec:numeric_learning}

Dual-Path Physiological Grounding extracts physiological evidence, but HR is still generated as discrete tokens under an objective that does not encode numerical distance, place value, or cross-sample ordering. Structured Numerical Alignment therefore introduces place-value-aware individual-sample supervision and cross-sample ordinal supervision, as shown in Fig.~\ref{fig:vital_overview}(b) and (c).

\subsubsection{Place-Value-Aware Numerical Supervision}

Given the ground-truth HR $h_i$ of sample $i$, we use a fixed formatting function $F(\cdot)$ to convert it into a standardized decimal string with consistent precision, $s_i=F(h_i)$. For example, when one decimal place is retained, $h_i=78.4$ corresponds to
$s_i=\texttt{``78.4''}$ and is embedded into the fixed-format target response
\texttt{``The estimated HR is 78.4 bpm.''}
We use tokenizer offsets to locate HR token positions $\Omega_i$ and supervise only positions with unambiguous digit correspondence. Inspired by Number Token Loss~\cite{zausinger2025regress} and NTIL's positional weighting~\cite{fei2025advancing}, we complement within-sample numerical supervision with cross-video HR ordering at the first differing digit.

For each valid digit position $t$, we normalize the logits over the digit tokens from $0$ to $9$ into conditional probabilities $q_{i,t}(k)$ and compute the expected digit error according to the distance between the predicted digit and the ground-truth digit $d^{*}_{i,t}$. Because errors at different decimal positions contribute differently to the final HR error, we further weight the digit error using the corresponding decimal place-value weight $w_{i,t}$:
\begin{equation}
\mathcal{L}_{\mathrm{PV}}
=
\frac{
\sum_{i}
\sum_{t \in \Omega_i}
w_{i,t} r_{i,t}
\sum_{k=0}^{9}
q_{i,t}(k)
\left|k-d^{*}_{i,t}\right|
}{
\sum_{i}
\sum_{t \in \Omega_i}
w_{i,t} r_{i,t}
+\epsilon
}
\label{eq:place_value_loss}
\end{equation}
where $\Omega_i$ denotes the set of numeric-token positions for sample $i$, $d^{*}_{i,t}$ is the ground-truth digit at position $t$, and $r_{i,t}$ indicates whether the position has a valid and unambiguous digit-token alignment. As a result, prediction errors at more significant decimal positions receive larger penalties than those at less significant positions. Details of character-span mapping, place-value weight normalization, and valid-token filtering are provided in Appendix~\ref{sec:supp_implementation}.

\subsubsection{Cross-Sample Ordinal Supervision}

Place-value-aware supervision improves the numerical fidelity of individual predictions but does not explicitly preserve HR ordering across videos. Within each mini-batch, we sort samples by ground-truth HR and construct candidate pairs using predefined rank offsets. We retain pairs \((i,j)\) satisfying \(h_i<h_j\), whose HR difference exceeds a noise threshold and whose formatted HR strings have the same number of integer digits. For each pair, we locate the first differing digit in the two standardized HR strings and identify the target tokens covering this digit. Pairs with incomparable token-span structures are excluded.

Let \(u^-_{ij}\) and \(u^+_{ij}\) denote the target tokens covering the first differing digit in the lower- and higher-HR strings, respectively, and let \(t_i^*\) and \(t_j^*\) denote their corresponding output positions. We define the token-preference margin between the two samples as
\begin{equation}
\begin{aligned}
M_{ij}
={}&
\left(
z_{j,t_j^{*},u^{+}_{ij}}
-
z_{j,t_j^{*},u^{-}_{ij}}
\right)
-
\left(
z_{i,t_i^{*},u^{+}_{ij}}
-
z_{i,t_i^{*},u^{-}_{ij}}
\right)
\end{aligned}
\label{eq:ordinal_margin}
\end{equation}
where $z_{i,t,u}$ denotes the logit assigned to vocabulary token $u$ at output position $t$ for sample $i$. Based on this preference margin, the cross-sample ordinal loss is defined as
\begin{equation}
\mathcal{L}_{\mathrm{ord}}
=
\frac{
\sum_{(i,j)\in\mathcal{P}}
\alpha_{ij}
\operatorname{softplus}
\left(-M_{ij}/\tau\right)
}{
\sum_{(i,j)\in\mathcal{P}}
\alpha_{ij}
+\epsilon
}
\label{eq:ordinal_loss}
\end{equation}

$\mathcal{P}$ is the set of valid sample pairs, $\tau$ is a temperature parameter, and $\alpha_{ij}$ adjusts the strength of the constraint according to the ground-truth HR difference between the two samples. This objective requires the video with the higher HR to exhibit a stronger preference for the higher-value token at the first differing digit, reducing the tendency for predictions from different samples to contract toward the same high-density region of the source-domain HR distribution. Details of sample-pair construction and token alignment are provided in Appendix~\ref{sec:supp_implementation}.

\subsection{Overall Objective}
\label{sec:overall_objective}

The complete training objective is
\begin{equation}
\mathcal{L}
=
\mathcal{L}_{\mathrm{LM}}
+\lambda_{\mathrm{PV}}\mathcal{L}_{\mathrm{PV}}
+\lambda_{\mathrm{ord}}\mathcal{L}_{\mathrm{ord}}
+\lambda_{\mathrm{aux}}\mathcal{L}_{\mathrm{aux}}
\label{eq:overall_loss}
\end{equation}

$\lambda_{\mathrm{PV}}$, $\lambda_{\mathrm{ord}}$, and $\lambda_{\mathrm{aux}}$ control the strengths of the place-value-aware numerical supervision, cross-sample ordinal supervision, and auxiliary physiological regression supervision, respectively.

Stage I uses only PhysTokens, whereas Stage II restores native semantic visual tokens and jointly optimizes Eq. (7). For general-purpose video instruction samples, only \(L_{\mathrm{LM}}\) is computed. At inference, the auxiliary regression head and additional training losses are removed, and HR is generated through the original language head.

\begin{table*}[t]
\centering
\begingroup
\small
\setlength{\tabcolsep}{2.5pt}
\begin{tabular}{@{}lcccccccccccccccc@{}}
\toprule
\multirow{2}{*}{Method}
& \multicolumn{4}{c}{PURE}
& \multicolumn{4}{c}{UBFC-rPPG}
& \multicolumn{4}{c}{BUAA}
& \multicolumn{4}{c}{VIPL-HR} \\
\cmidrule(lr){2-5}
\cmidrule(lr){6-9}
\cmidrule(lr){10-13}
\cmidrule(lr){14-17}
& MAE & RMSE & $r$ & SD
& MAE & RMSE & $r$ & SD
& MAE & RMSE & $r$ & SD
& MAE & RMSE & $r$ & SD \\
\midrule
Green  & 15.47  & 23.31  & 0.23 & 22.29 & 12.77 & 15.88 & 0.29 & 13.72 & 8.85 & 13.54 & 0.44 & 9.84 & 13.67 & 29.38 & 0.10 & 22.00 \\
CHROM  & 9.26 & 14.12 & 0.73 & 13.97 & 9.60 & 13.87 & 0.43 & 11.56 & 6.58 & 11.88 & 0.55 & 8.72 & 18.87 & 24.28 & 0.24 & 20.19 \\
POS & 9.58 & 16.27 & 0.65 & 15.55 & 9.03 & 11.20 & 0.40 & 9.28 & 7.20 & 12.54 & 0.56 & 8.71 & 19.18 & 24.93 & 0.23 & 20.33 \\
\midrule
PhysNet & 10.96 & 15.88 & 0.54 & 13.01 & 8.09 & 13.26 & 0.68 & 11.25 & 6.04 & 8.36 & 0.56 & 8.31 & 14.81 & 18.62 & 0.17 & 18.44 \\
PhysFormer & 17.51 & 21.45 & 0.14 & 19.96 & 14.88 & 18.72 & 0.26 & 15.08 & 8.81 & 11.82 & 0.49 & 11.78 & 14.00 & 17.39 & 0.28 & 17.15 \\
RhythmFormer & 9.06 & 13.97 & 0.62 & 12.16 & 6.55 & 10.48 & 0.72 & 10.06 & 5.95  & 8.13 & 0.61 & 7.84 & 11.81 & 17.02 & 0.36 & 16.28 \\
HSRD$^\dagger$ & 6.47 & 11.03 & 0.81 & 9.21 & 4.78 & 7.31 & 0.89 & 7.07 & \underline{3.29} & 5.82 & 0.87 & 5.40 & 9.25 & 14.85 & 0.60 & 14.13 \\
RhythmGaussian$^\dagger$ & 7.51 & 10.90 & 0.77 & 10.53 & 5.11 & 7.62 & 0.79 & 7.45 & 3.36 & \textbf{4.94}$^*$ & \textbf{0.91} & 4.88 & 10.45 & 14.03 & 0.49 & 13.91 \\
FactorizePhys & 4.44  & \textbf{8.15}$^*$  & \textbf{0.92}$^*$  & \textbf{7.93}$^*$  & 3.75  & 5.60 & 0.92  & 5.49  & \textbf{3.05}$^*$  & \underline{4.96} & \underline{0.89} & 4.85 & 9.47  & 13.04  & 0.63  & 12.89 \\
\midrule
Period-LLM & 18.32  & 26.03  & 0.13  & 24.50  & 16.57  & 23.18 & 0.21  & 22.34  & 15.88  & 24.58 & 0.30 & 22.97 & 20.79  & 25.03  & 0.09  & 23.08 \\
PhysLLM & 6.95  & 10.66  & 0.85  & 10.17  & 4.17  & 7.07 & 0.84  & 6.65  & 7.05  & 8.39 & 0.85 & 7.61 & 10.32  & 13.49  & 0.56  & 12.29 \\
\midrule

w/o Semantic Path & 4.99  & 11.16 & 0.83  & 11.05  & 2.62 & 4.88 & 0.95 & 4.47 & 6.97 & 8.32 & 0.84 &  6.22 & 9.17 & 13.58 & 0.63 & 13.05\\
w/o SNA & 4.83  & 10.87 & 0.85  & 10.55  & 2.39 & 4.12 & 0.96 & 4.03 & 6.38 & 8.15 & 0.87 &  5.29 & 8.90 & 13.11 & 0.67 & 12.18\\

w/o $\mathcal{L}_{\mathrm{ord}}$ & 4.33 & 9.54 & \underline{0.89} & 9.32 & \underline{2.26} & 3.97 & 0.95 & 3.89 & 5.42 & 7.32 & 0.88 & 5.19 & \underline{8.67} & \underline{12.20} & 0.67 & 12.19 \\
w/o $\mathcal{L}_{\mathrm{PV}}$ & 4.40 & 9.89 & 0.85 & 9.51 & 2.41 & \underline{3.86} & \textbf{0.98} & \underline{3.60} & 4.73 & 6.76 & \underline{0.89} & \underline{4.64} & 8.73 & 12.59 & \underline{0.69} & \underline{11.84} \\
w/o $\mathcal{L}_{\mathrm{aux}}$ & \underline{4.04}  & 9.38 & 0.87  & 9.31  & 2.51 & 3.95 & \underline{0.97} & 3.76 & 5.03 & 7.41 & 0.88 & 5.34 & 9.12 & 13.05 & 0.66 & 12.04\\
VITAL & \textbf{3.97}$^*$  & \underline{9.29} & \underline{0.89}  & \underline{9.15}  & \textbf{2.14}$^*$  & \textbf{3.64}$^*$ & \textbf{0.98}$^*$ & \textbf{3.42}$^*$ &4.70  &6.40  &\textbf{0.91}$^*$  & \textbf{4.56}$^*$ & \textbf{8.61}$^*$  & \textbf{11.95}$^*$ & \textbf{0.71}$^*$ & \textbf{11.61}$^*$ \\
\bottomrule
\end{tabular}
\endgroup
\caption{HR estimation results on MSDG protocol. Lower MAE/RMSE/SD and higher $r$ are better; w/o denotes VITAL ablations. $^\dagger$ denotes STMap-based methods. In this and following tables, best results are bolded, and second-best results are underlined. Significant ($p$<.05) improvements are marked '$^*$' using paired $t$-tests.}
\label{tab:MSDG}
\end{table*}

\section{Experiments}
\label{sec:experiments}

\subsection{Datasets and Baselines}

We evaluate our method on six rPPG datasets covering diverse recording conditions, HR distributions~\cite{yang2025fedhug}, and participant characteristics. UBFC-rPPG \cite{bobbia2019unsupervised} contains 42 facial videos recorded under sunlight and indoor illumination. BUAA \cite{xi2020image} contains facial videos collected under multiple lighting conditions. PURE \cite{stricker2014non} consists of 60 RGB videos from 10 participants performing six activities, including sitting still and head rotation. VIPL-HR \cite{niu2018vipl} contains facial videos captured using three RGB cameras under different illumination conditions and motion levels. MMPD \cite{tang2023mmpd} contains a diverse collection of videos covering different skin tones, physical conditions, and motion states. PhysDrive \cite{wang2026physdrive} is a multimodal dataset collected in natural driving environments with varying illumination and facial motion conditions. MMPD and PhysDrive are treated as challenging out-of-domain datasets, and the corresponding results are reported in Tab.~\ref{tab:hard_datasets}.

For a comprehensive comparison, we include conventional signal-processing methods, including Green \cite{verkruysse2008remote}, CHROM \cite{de2013robust}, and POS \cite{wang2016algorithmic}; deep learning methods based on STMaps \cite{niu2019rhythmnet} or facial videos, including PhysNet \cite{yu2019remote}, PhysFormer \cite{yu2022physformer}, HSRD \cite{wang2023hierarchical}, RhythmFormer \cite{zou2025rhythmformer}, RhythmGaussian \cite{lu2025rhythmguassian} and FactorizePhys \cite{joshi2024factorizephys}; and large-model-based methods, including Period-LLM \cite{zhang2025period} and PhysLLM \cite{xie2025physllm}.

\subsection{Implementation Details}

\textbf{Training Details.}
We use Qwen3-VL-4B-Instruct as the base VLM and FactorizePhys as the physiological encoder. For each MSDG split, it is initialized from a FactorizePhys model trained only on the corresponding source domains. Each video is sampled to 160 frames, resized to \(72\times72\), and converted into 159 frame-difference features, which are mapped into PhysTokens by a two-layer MLP. During VLM adaptation, both visual encoders and the auxiliary regression head remain frozen, while the physiological projector and language-model LoRA parameters are optimized. After projector warm-up, Stage I uses only PhysTokens, whereas Stage II introduces native visual tokens and mixes rPPG and ActivityNet-QA samples at a \(4{:}1\) ratio. We set \(\lambda_{\mathrm{PV}}=0.5\), \(\lambda_{\mathrm{ord}}=0.1\), and \(\lambda_{\mathrm{aux}}=0.1\). More details are provided in Appendix~\ref{sec:supp_implementation}.

\textbf{Performance Metrics.}
We report the average standard deviation of the error (SD), mean absolute error (MAE), root mean square error (RMSE), and Pearson’s correlation coefficient ($r$) of five runs with different random seeds.

\subsection{Multi-Source Domain Generalization}

\subsubsection{HR Estimation}

We first evaluate cross-dataset HR estimation under the standard multi-source domain generalization (MSDG) protocol~\cite{wang2025dgrppg}. Specifically, three datasets from PURE, UBFC-rPPG, BUAA, and VIPL-HR are used as source domains, while the remaining dataset serves as the unseen target domain \cite{lu2023neuron}. No target-domain samples are used for training or model selection. The results are in Tab.~\ref{tab:MSDG}.

VITAL achieves the lowest MAE on three of four target domains and the best overall results on UBFC-rPPG and VIPL-HR. It consistently outperforms Period-LLM and PhysLLM. Despite sharing FactorizePhys's physiological encoder, VITAL achieves lower MAE on three targets, indicating benefits beyond encoder choice. On BUAA, it achieves the highest \(r\) and lowest SD, indicating more reliable tracking of target-domain variation.

\subsubsection{Ablation Study}
We evaluate the contributions of the native semantic pathway, Structured Numerical Alignment, and auxiliary regression supervision under the MSDG protocol. As shown in Table 2, removing the semantic pathway consistently degrades performance, with the largest deterioration observed on BUAA, supporting the benefit of retaining semantic visual evidence alongside PhysTokens. Removing SNA also worsens performance across all target domains. Removing \(L_{\mathrm{PV}}\) or \(L_{\mathrm{ord}}\) individually yields intermediate results, confirming that place-value-aware supervision and cross-sample ordinal supervision play complementary roles. Performance also decreases without \(L_{\mathrm{aux}}\), supporting its role in preserving decodable continuous HR information in PhysTokens. Overall, the full VITAL provides the most consistent cross-domain performance.

\subsubsection{Generalization to Challenging Datasets}

To further evaluate generalization under more complex real-world conditions, we train the models on the combined training sets of PURE, UBFC-rPPG, BUAA, and VIPL-HR, and directly evaluate them on PhysDrive and MMPD without using either dataset during training or model selection. Compared with the preceding MSDG setting, this evaluation involves both larger cross-dataset distribution shifts and more challenging acquisition conditions. PhysDrive was collected in real driving environments and includes complex illumination changes, driver motion, and natural driving conditions. MMPD covers different skin tones, physical conditions, and motion states, resulting in substantial variation across participants and recording conditions. Both datasets therefore differ considerably from the training data. Tab.~\ref{tab:hard_datasets} reports the HR estimation results on these two challenging datasets. VITAL maintains favorable performance under substantial distribution shifts, indicating that its joint modeling of physiological evidence and language representations improves generalization in challenging scenarios.

\begin{table}[t]
\centering
\begingroup
\small
\setlength{\tabcolsep}{1.8pt}
\begin{tabular}{@{}llcccc@{}}
\toprule
Method & Dataset & MAE$\downarrow$ & RMSE$\downarrow$ & $r\uparrow$ & SD$\downarrow$ \\
\midrule
\multirow{2}{*}{FactorizePhys} & PhysDrive & \underline{14.29} & \underline{17.88} & \underline{0.13} & \underline{15.96} \\
 & MMPD & 12.74 & \underline{15.04} & \underline{0.17} & \underline{13.37} \\
\multirow{2}{*}{Period-LLM} & PhysDrive & 24.04 & 26.39 & 0.01 & 23.61 \\
 & MMPD & 23.42 & 28.97 & 0.02 & 26.55 \\
\multirow{2}{*}{PhysLLM} & PhysDrive & 16.63 & 19.32 & 0.11 & 16.90 \\
 & MMPD & \underline{12.49} & 15.72 & \underline{0.17} & 14.75 \\
\multirow{2}{*}{VITAL} & PhysDrive & \textbf{12.65}$^*$ & \textbf{15.08}$^*$ & \textbf{0.17}$^*$ & \textbf{13.74}$^*$ \\
 & MMPD & \textbf{10.63}$^*$ & \textbf{13.23}$^*$ & \textbf{0.22}$^*$ & \textbf{12.90}$^*$ \\
\bottomrule
\end{tabular}
\endgroup
\caption{HR estimation results on challenging datasets.}
\label{tab:hard_datasets}
\end{table}

\subsubsection{Visualization Analysis}

\begin{figure}[t]
    \centering
    \includegraphics[width=\columnwidth]{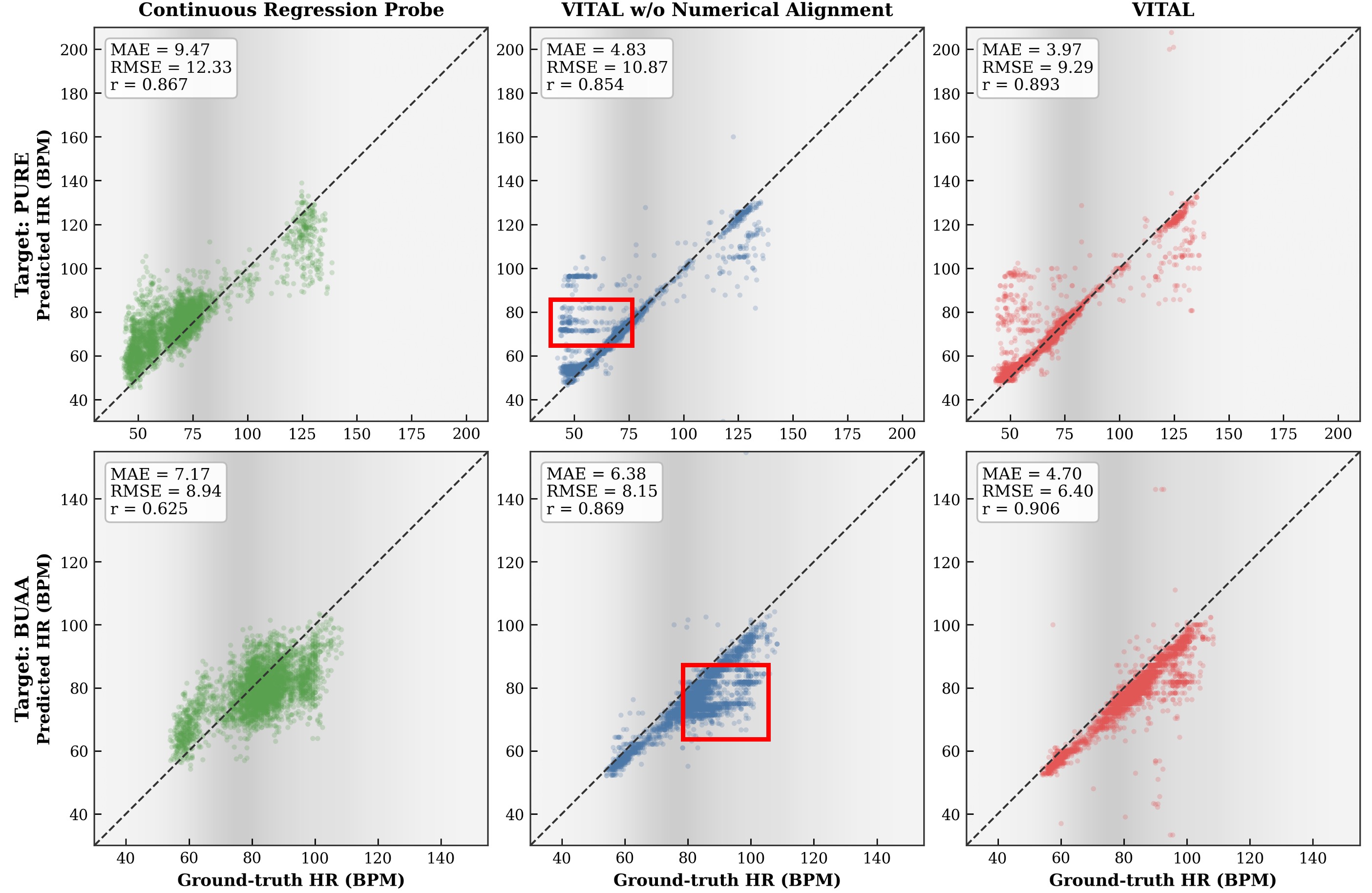}
    \caption{Prediction distributions on PURE and BUAA. Red boxes highlight repeated-value banding.}
    \label{fig:prediction_vis}
\end{figure}

As shown in Fig.~\ref{fig:prediction_vis}, we use PURE and BUAA as representative examples. The continuous regression probe directly decodes HR from the physiological representation and produces relatively smooth predictions despite cross-domain contraction, indicating that continuous HR information remains decodable. Without Structured Numerical Alignment, autoregressive predictions exhibit clear horizontal banding, where samples with different ground-truth HR values are mapped to the same or similar outputs. After introducing Structured Numerical Alignment, this concentration is substantially reduced and predictions vary more continuously with ground-truth HR, indicating that the proposed alignment effectively mitigates source-prior attraction. See Supplementary Table~12 for quantitative results.

\subsection{Further Analysis}

\subsubsection{General Capability and Efficiency.}
Tab.~\ref{tab:general_efficiency} compares video understanding, zero-shot MMPD performance, and model overhead. All Period-LLM runs use Qwen3-VL-4B, with HR-only fine-tuning in Tables~\ref{tab:MSDG} and~\ref{tab:hard_datasets} and $4{:}1$ rPPG-to-QA mixing here. VITAL scores 60.53\% on MVBench, 3.52 percentage points below the native VLM and 1.15 above Period-LLM. Its overhead over the native VLM is 22M parameters and 0.236 GiB peak memory.

\begin{table}[!t]
\centering
\begingroup
\small
\setlength{\tabcolsep}{2pt}
\begin{tabular}{@{}lcccc@{}}
\toprule
Method & \shortstack{MVBench$\uparrow$\\(\%)} & MAE$\downarrow$ & \shortstack{Params$\downarrow$\\(B)} & \shortstack{Memory$\downarrow$\\(GiB)} \\
\midrule
Qwen3VL(native) & \textbf{64.05}$^*$ & 20.98 & \textbf{4.438} & \textbf{8.512} \\
Period-LLM & 59.38 & \underline{17.89} & \textbf{4.438} & \textbf{8.512} \\
VITAL & \underline{60.53} & \textbf{10.63}$^*$ & \underline{4.460} & \underline{8.748} \\
\bottomrule
\end{tabular}
\endgroup
\caption{General video understanding on MVBench, physiological performance on MMPD, and model overhead with Qwen3-VL-4B backbone.}
\label{tab:general_efficiency}
\end{table}

\subsubsection{Generalization across VLM Backbones}

\looseness=-1 To evaluate backbone generalization, we apply VITAL to Qwen3-VL-2B, 4B, and 8B, as well as the architecturally different LLaVA-NeXT-Video-7B~\cite{li2024llava}, while keeping the physiological encoder and training protocol unchanged. Following the MSDG setting, models are trained on three source domains and evaluated on PURE or UBFC-rPPG as the unseen target domain. As shown in Tab. ~\ref{tab:backbone_generalization}, performance generally improves with Qwen3-VL model scale, while VITAL also performs consistently with LLaVA-NeXT-Video-7B, demonstrating scalability across model sizes and generalization across VLM architectures.

\begin{table}[!t]
\centering
\begingroup
\small
\setlength{\tabcolsep}{1.8pt}
\begin{tabular}{@{}llcccc@{}}
\toprule
Method & Dataset & MAE$\downarrow$ & RMSE$\downarrow$ & $r\uparrow$ & SD$\downarrow$ \\
\midrule
\multirow{2}{*}{Qwen3-VL-2B} & PURE & 4.51 & 10.67 & 0.85 & 10.47 \\
 & UBFC-rPPG & 2.98 & 5.63 & 0.95 & 5.25 \\
\multirow{2}{*}{Qwen3-VL-4B} & PURE & \underline{3.97} & \underline{9.29} & \underline{0.89} & \underline{9.15} \\
 & UBFC-rPPG & \underline{2.14} & \underline{3.64} & \textbf{0.98} & \underline{3.42} \\
\multirow{2}{*}{Qwen3-VL-8B} & PURE & \textbf{3.88}$^*$ & \textbf{9.14}$^*$ & \textbf{0.91}$^*$ & \textbf{9.03}$^*$ \\
 & UBFC-rPPG & \textbf{2.11}$^*$ & \textbf{3.12}$^*$ & \textbf{0.98}$^*$ & \textbf{2.96}$^*$ \\
\multirow{2}{*}{\shortstack[l]{LLaVA-NeXT-\\Video-7B}} & PURE & 4.86 & 9.57 & 0.88 & 9.45 \\
 & UBFC-rPPG & 2.77 & 4.12 & \underline{0.97} & 3.48 \\
\bottomrule
\end{tabular}
\endgroup
\caption{Cross-domain HR estimation of VITAL with different VLM backbones under the MSDG protocol. Models are trained on the other three source domains and evaluated on the indicated unseen target domain.}
\label{tab:backbone_generalization}
\end{table}

\subsubsection{Joint Evaluation of Semantic and Physiological States}

To evaluate whether VITAL retains visual semantic capabilities while leveraging physiological evidence for human-state understanding, we conduct zero-shot evaluation on the held-out MMPD dataset. We construct subsets for glasses-wearing recognition, LED-low/LED-high lighting recognition, resting versus post-exercise recovery classification, and HR estimation. For recovery classification, only stationary videos are retained to minimize explicit motion cues.

All tasks are jointly evaluated using a single multi-question prompt. All models receive the same semantic video frames, while VITAL additionally uses PhysTokens extracted from the full sequence. Each model generates a unified response covering glasses wearing, lighting condition, recovery state, and HR. As shown in Tab.~\ref{tab:mmpd_multitask}, VITAL maintains broadly comparable visual recognition performance while improving recovery-state classification. Since resting and post-exercise recovery can appear visually similar but differ physiologically, this result suggests that physiological measurement provides complementary evidence for visually ambiguous human-state inference, extending RPM beyond numerical measurement.

\begin{table}[t]
\centering
\begingroup
\small
\setlength{\tabcolsep}{2pt}
\begin{tabular}{@{}lccccc@{}}
\toprule
\multirow{2}{*}{Method}
& \multicolumn{3}{c}{Balanced accuracy (\%) $\uparrow$}
& \multicolumn{2}{c}{HR Estimation} \\
\cmidrule(lr){2-4}
\cmidrule(lr){5-6}
& Glasses
& Lighting
& Recovery
& MAE $\downarrow$
& $r$ $\uparrow$ \\
\midrule
Qwen3VL(native)
& \textbf{89.55\%}$^*$ & \underline{46.94\%}  & 50.04\% & 21.76 & 0.01  \\
Period-LLM
& 83.55\% & 45.10\%  & \underline{50.16\%} & \underline{18.69} & \underline{0.03} \\
VITAL
& \underline{84.59\%} & \textbf{49.81\%}$^*$ & \textbf{56.54\%}$^*$ & \textbf{10.91}$^*$ & \textbf{0.18}$^*$  \\
\bottomrule
\end{tabular}
\endgroup
\caption{Zero-shot semantic and physiological evaluation on MMPD using one multi-question prompt.}
\label{tab:mmpd_multitask}
\end{table}

\section{Conclusion}

This work identifies two bottlenecks in adapting general-purpose VLMs to remote HR estimation: semantic visual representations may discard fine-grained physiological cues, while token-wise generation is poorly aligned with continuous numerical measurement. VITAL addresses them through a parallel physiological pathway and Structured Numerical Alignment. Experiments across multiple rPPG datasets improve cross-domain HR estimation and reduce source-prior attraction, while limiting degradation in video understanding and preserving the original language-generation interface. Future work will pursue capability-preserving adaptation, reduce reliance on dedicated physiological encoders, and extend VITAL beyond HR to multiple physiological signals.

\clearpage
\appendix
\setcounter{dbltopnumber}{3}
\renewcommand{\dbltopfraction}{0.85}
\setcounter{secnumdepth}{1}
\setcounter{table}{0}
\setcounter{figure}{0}
\setcounter{equation}{0}
\renewcommand{\thetable}{S\arabic{table}}
\renewcommand{\thefigure}{S\arabic{figure}}
\renewcommand{\theequation}{S\arabic{equation}}
\twocolumn[{\centering\LARGE\bfseries Supplementary Material\par\medskip}]
\input{supplement}

\bibliography{aaai2027}
\end{document}

%% file: supplement.tex
\section{Additional Implementation Details}
\label{sec:supp_implementation}

\subsection{Optimization Settings}

Projector warm-up, Stage I, and Stage II are trained for
60, 20, and 10 epochs, respectively.
We use AdamW with a learning rate of
$5\times10^{-4}$ for projector warm-up and
$1\times10^{-4}$ for jointly optimizing the physiological
projector and language-model LoRA parameters in Stages I and II.
The projector warm-up uses a batch size of 32. Stages I and II
use a per-device batch size of 4 with four-step gradient
accumulation, resulting in an effective batch size of 16.
Unless otherwise specified, the physiological feature sequence is
temporally pooled into 20 PhysTokens for both training and inference.
LoRA is applied to all linear layers in the language model,
excluding the visual encoder, with rank 8, scaling factor
16, and dropout 0.05.

\subsection{Character-to-Token Alignment}

Ground-truth HR values are formatted with one decimal place and
embedded into a fixed assistant-response template. We tokenize the
normalized HR string with offset mappings and match its token-ID
sequence exactly to the non-masked assistant target tokens in the
complete templated response.

An aligned numeric-token position is considered valid for
Place-Value-Aware Numerical Supervision only when its character span
contains exactly one digit and its corresponding shifted
language-model position
$t=\mathrm{label\_position}-1$ lies within the valid logit range.
Tokens whose spans contain multiple digits, tokens without valid
non-empty offset spans, and tokens whose shifted logit positions are
out of range are excluded from $\mathcal{L}_{\mathrm{PV}}$.
These excluded tokens remain supervised by the standard
language-modeling objective $\mathcal{L}_{\mathrm{LM}}$.
Decimal points and non-numeric characters are not included in
$\mathcal{L}_{\mathrm{PV}}$.

\subsection{Place-Value Weighting}

Let $\mathcal{V}_i$ denote the set of valid single-digit
numeric-token positions in sample $i$. For each
$t\in\mathcal{V}_i$, let $p_{i,t}$ denote the decimal position of the
unique digit covered by token position $t$. Its unnormalized
place-value weight is defined as

\begin{equation}
\bar{w}_{i,t}
=
10^{\beta p_{i,t}},
\qquad \beta = 0.5.
\label{eq:supp_place_weight}
\end{equation}

where $p_{i,t}=2,1,0,-1$ correspond to the hundreds, tens, units,
and tenths positions, respectively. The weights are normalized as

\begin{equation}
w_{i,t}
=
\operatorname{clip}
\left(
\frac{\bar{w}_{i,t}}
{\frac{1}{|\mathcal{V}_i|}
\sum_{u\in\mathcal{V}_i}\bar{w}_{i,u}},
\,0.25,\,4.0
\right).
\label{eq:supp_place_weight_normalization}
\end{equation}

This normalization gives the valid numeric-token weights within each
sample a unit mean before clipping, thereby controlling the overall
loss scale across samples. The clipping operation limits the influence
of individual positions, while the exponential place-value function
assigns larger penalties to errors at more significant decimal
positions. The final place-value-aware loss is computed as

\begin{equation}
\mathcal{L}_{\mathrm{PV}}
=
\frac{
\sum_i\sum_{t\in\mathcal{V}_i}
w_{i,t}\ell_{i,t}^{\mathrm{digit}}
}{
\sum_i\sum_{t\in\mathcal{V}_i}w_{i,t}+\epsilon
},
\label{eq:supp_place_value_loss}
\end{equation}

where $\ell_{i,t}^{\mathrm{digit}}$ is the expected digit-distance
loss defined in the main paper.

\subsection{Ordinal pair construction}
Within each mini-batch, samples are sorted by ground-truth HR, and
initial candidate pairs are constructed using rank offsets
$\{1,2,4,8\}$. The pairs are then filtered sequentially according to
the following criteria: 1) the HR difference strictly exceeds
$\Delta_{\mathrm{HR}}=3$ BPM; 2) the two formatted HR strings have the
same number of integer digits; and 3) the first differing digits have
compatible token-span structures.

For each retained pair, the first differing digit is located from the
most significant to the least significant position. Integer digits are
examined first, followed by fractional digits. The corresponding
lower- and higher-value digit tokens define $u_{ij}^{-}$ and
$u_{ij}^{+}$, respectively. Let $\operatorname{shape}(x)$ replace every
digit character in token substring $x$ with the symbol $\mathrm{D}$
while retaining all non-digit characters. Two token spans are
considered compatible when

\begin{equation}
\operatorname{shape}(x_{ij}^{-})
=
\operatorname{shape}(x_{ij}^{+}),
\end{equation}

where $x_{ij}^{-}$ and $x_{ij}^{+}$ are the substrings covered by the
two tokens. The two selected positions must also be enabled by the
numerical-loss mask, and their token IDs must be valid and distinct.
Pairs with missing covering tokens, incompatible span structures,
masked positions, out-of-range token IDs, or identical token IDs are
discarded. If no valid pair remains in a mini-batch,
$\mathcal{L}_{\mathrm{ord}}$ is set to a differentiable zero.

The pair weight used in $\mathcal{L}_{\mathrm{ord}}$ is

\begin{equation}
\alpha_{ij}
=
\min\left(
1,\,
\frac{h_j-h_i}{15}
\right),
\qquad h_j>h_i.
\label{eq:supp_pair_weight}
\end{equation}

and the temperature is set to $\tau=1$.

\begin{table*}[t]
\centering
\begingroup
\small
\setlength{\tabcolsep}{2pt}
\begin{tabular}{@{}lccc*{8}{c}@{}}
\toprule
\multirow{2}{*}{Training Schedule} & \multirow{2}{*}{Warm-up} & \multirow{2}{*}{Stage I} & \multirow{2}{*}{Stage II}
& \multicolumn{4}{c}{PURE} & \multicolumn{4}{c}{UBFC-rPPG} \\
\cmidrule(lr){5-8}\cmidrule(lr){9-12}
& & & & MAE$\downarrow$ & RMSE$\downarrow$ & $r\uparrow$ & SD$\downarrow$ & MAE$\downarrow$ & RMSE$\downarrow$ & $r\uparrow$ & SD$\downarrow$ \\
\midrule
Stage II & No & No & Yes & 14.21 & 16.98 & 0.62 & 15.07 & 7.45 & 10.19 & 0.73 & 8.58 \\
Warm-up $\rightarrow$ Stage II & Yes & No & Yes & 13.49 & 16.74 & 0.65 & 14.16 & 7.83 & 10.32 & 0.74 & 8.59 \\
Stage I $\rightarrow$ Stage II & No & Yes & Yes & \underline{4.35} & \underline{10.27} & \underline{0.87} & \underline{10.03} & \underline{2.71} & \underline{4.15} & \underline{0.97} & \underline{3.80} \\
Warm-up $\rightarrow$ Stage I $\rightarrow$ Stage II & Yes & Yes & Yes & \textbf{3.97} & \textbf{9.29} & \textbf{0.89} & \textbf{9.15} & \textbf{2.14} & \textbf{3.64} & \textbf{0.98} & \textbf{3.42} \\
\bottomrule
\end{tabular}
\par\medskip
\begin{tabular}{@{}lccc*{8}{c}@{}}
\toprule
\multirow{2}{*}{Training Schedule} & \multirow{2}{*}{Warm-up} & \multirow{2}{*}{Stage I} & \multirow{2}{*}{Stage II}
& \multicolumn{4}{c}{BUAA} & \multicolumn{4}{c}{VIPL-HR} \\
\cmidrule(lr){5-8}\cmidrule(lr){9-12}
& & & & MAE$\downarrow$ & RMSE$\downarrow$ & $r\uparrow$ & SD$\downarrow$ & MAE$\downarrow$ & RMSE$\downarrow$ & $r\uparrow$ & SD$\downarrow$ \\
\midrule
Stage II & No & No & Yes & 15.03 & 18.74 & 0.72 & 16.62 & 18.24 & 22.06 & 0.51 & 19.35 \\
Warm-up $\rightarrow$ Stage II & Yes & No & Yes & 14.76 & 17.09 & 0.75 & 15.95 & 16.43 & 21.21 & 0.55 & 18.89 \\
Stage I $\rightarrow$ Stage II & No & Yes & Yes & \underline{4.98} & \underline{6.81} & \underline{0.89} & \underline{4.72} & \underline{8.64} & \underline{12.13} & \underline{0.70} & \underline{11.88} \\
Warm-up $\rightarrow$ Stage I $\rightarrow$ Stage II & Yes & Yes & Yes & \textbf{4.70} & \textbf{6.40} & \textbf{0.91} & \textbf{4.56} & \textbf{8.61} & \textbf{11.95} & \textbf{0.71} & \textbf{11.61} \\
\bottomrule
\end{tabular}
\endgroup
\caption{Ablation of projector warm-up and the two stages of
Progressive PhysToken Alignment under the MSDG protocol. Each dataset
serves once as the unseen target domain, with the remaining three
datasets used as source domains. All schedules use the same total
optimization budget and matched numbers of rPPG and general-purpose
instruction samples. The best and second-best results are shown in
bold and underlined, respectively.}
\label{tab:supp_progressive_stages}
\end{table*}

\section{Additional Experimental Analysis}
\label{sec:supp_additional_analysis}

\subsection{Ablation of the Progressive PhysToken Alignment Schedule}
\label{sec:supp_progressive_alignment}

The complete alignment schedule comprises an
auxiliary-regression-based projector warm-up followed by the two
stages of Progressive PhysToken Alignment: PhysToken-only alignment
in Stage I and dual-path adaptation in Stage II. To isolate the
contribution of each component, we compare four training schedules
under the MSDG protocol. \textbf{Stage II} directly performs dual-path
adaptation without projector warm-up or Stage I.
\textbf{Warm-up $\rightarrow$ Stage II} initializes the physiological
projector through auxiliary HR regression before dual-path adaptation.
\textbf{Stage I $\rightarrow$ Stage II} retains PhysToken-only
alignment but omits projector warm-up. The \textbf{full schedule}
applies all three components sequentially.

Each of PURE, UBFC-rPPG, BUAA, and VIPL-HR serves once as the unseen
target domain, with the remaining three datasets used as source
domains. Across schedules, we keep the model initialization, source
data, Structured Numerical Alignment, learning rates, LoRA
configurations, and total number of optimization steps fixed. When
projector warm-up or Stage I is removed, its optimization steps are
reassigned to the subsequent active stage. We further match the total
numbers of rPPG and general-purpose instruction samples across
schedules.

As shown in Table~\ref{tab:supp_progressive_stages}, directly entering
Stage II yields weak cross-domain performance on all four target
domains. Projector warm-up alone provides only limited and
dataset-dependent gains. Relative to direct Stage-II adaptation,
warm-up improves all four metrics on PURE, BUAA, and VIPL-HR, but
slightly increases MAE, RMSE, and SD on UBFC-rPPG despite a marginal
improvement in correlation. These results indicate that preserving
continuously decodable HR information in the physiological projector
is not, by itself, sufficient to establish a reliable mapping from
PhysTokens to autoregressive HR outputs.

Stage I provides the largest performance improvement. Adding
PhysToken-only alignment before Stage II reduces MAE from 14.21 to
4.35 on PURE, from 7.45 to 2.71 on UBFC-rPPG, from 15.03 to 4.98 on
BUAA, and from 18.24 to 8.64 on VIPL-HR. Correlation also improves
consistently across the four target domains. These results support the
proposed training order: the model should first associate
physiological representations with textual HR outputs before the
pretrained semantic pathway is introduced.

Adding projector warm-up before Stage I further improves all
dataset--metric combinations. The full schedule achieves MAEs of
3.97, 2.14, 4.70, and 8.61 on PURE, UBFC-rPPG, BUAA, and VIPL-HR,
respectively. The results suggest that warm-up provides a more suitable
initialization for subsequent PhysToken-only alignment, although
warm-up alone cannot replace Stage I. Because the optimization budget
and training-sample counts are matched across schedules, these
controlled comparisons support the benefit of progressively organizing
physiological grounding rather than simply increasing optimization
steps or data exposure.

\subsection{Comparison with Alternative Numerical Supervision Objectives}
\label{sec:supp_numerical_supervision}

We next examine whether the gains from Structured Numerical Alignment
can be reproduced by alternative numerical supervision objectives.
The comparison follows the same MSDG protocol, with PURE, UBFC-rPPG,
BUAA, and VIPL-HR serving in turn as the unseen target domain. All
variants use the same dual-path architecture, Progressive PhysToken
Alignment schedule, training data, optimization budget, and
hyperparameter settings.

For rPPG samples, all variants retain the language-modeling objective
$\mathcal{L}_{\mathrm{LM}}$ and auxiliary physiological regression
objective $\mathcal{L}_{\mathrm{aux}}$. Their shared objective is

\begin{equation}
\mathcal{L}_{\mathrm{base}}
=
\mathcal{L}_{\mathrm{LM}}
+
\lambda_{\mathrm{aux}}\mathcal{L}_{\mathrm{aux}}.
\label{eq:supp_base_objective}
\end{equation}

The variants differ only in the additional numerical supervision.
Number Token Loss~\cite{zausinger2025regress} introduces
regression-like supervision based on the numerical distance between
predicted and target number tokens, but does not distinguish errors
according to decimal significance. We further combine Number Token
Loss $\mathcal{L}_{\mathrm{NTL}}$ with Cross-Sample Ordinal
Supervision $\mathcal{L}_{\mathrm{ord}}$ to evaluate whether explicit
cross-sample ordering complements token-distance supervision.
Place-Value-Aware Numerical Supervision
$\mathcal{L}_{\mathrm{PV}}$ instead weights digit errors according to
their contribution to the final numerical value. The complete
Structured Numerical Alignment combines
$\mathcal{L}_{\mathrm{PV}}$ and $\mathcal{L}_{\mathrm{ord}}$. The
ordinal-only variant is reported as VITAL without
$\mathcal{L}_{\mathrm{PV}}$ in the main-paper ablation and is
therefore not duplicated here.

\begin{table*}[t]
\centering
\begingroup
\small
\setlength{\tabcolsep}{2pt}
\begin{tabular}{@{}l*{16}{c}@{}}
\toprule
\multirow{2}{*}{Training Objective}
& \multicolumn{4}{c}{PURE}
& \multicolumn{4}{c}{UBFC-rPPG}
& \multicolumn{4}{c}{BUAA}
& \multicolumn{4}{c}{VIPL-HR} \\
\cmidrule(lr){2-5}
\cmidrule(lr){6-9}
\cmidrule(lr){10-13}
\cmidrule(lr){14-17}
& MAE$\downarrow$
& RMSE$\downarrow$
& $r\uparrow$
& SD$\downarrow$
& MAE$\downarrow$
& RMSE$\downarrow$
& $r\uparrow$
& SD$\downarrow$
& MAE$\downarrow$
& RMSE$\downarrow$
& $r\uparrow$
& SD$\downarrow$
& MAE$\downarrow$
& RMSE$\downarrow$
& $r\uparrow$
& SD$\downarrow$ \\
\midrule

$\mathcal{L}_{\mathrm{base}}$
& 4.83
& 10.87
& 0.85
& 10.55
& 2.39
& 4.12
& \underline{0.96}
& 4.03
& 6.38
& 8.15
& 0.87
& 5.29
& 8.90
& 13.11
& 0.67
& 12.18 \\

$\mathcal{L}_{\mathrm{base}}
+\mathcal{L}_{\mathrm{NTL}}$
& 5.23
& 11.45
& 0.82
& 10.72
& 3.40
& 5.06
& 0.92
& 4.57
& 6.29
& 8.03
& \underline{0.88}
& 5.25
& 9.02
& 12.94
& \underline{0.68}
& \underline{12.05} \\

$\mathcal{L}_{\mathrm{base}}
+\mathcal{L}_{\mathrm{NTL}}
+\mathcal{L}_{\mathrm{ord}}$
& 4.35
& 9.76
& \underline{0.86}
& 9.63
& 2.61
& 4.53
& 0.95
& 4.39
& 6.01
& 7.83
& 0.87
& 5.20
& 8.72
& 12.49
& 0.67
& 12.17 \\

$\mathcal{L}_{\mathrm{base}}
+\mathcal{L}_{\mathrm{PV}}$
& \underline{4.33}
& \underline{9.54}
& \textbf{0.89}
& \underline{9.32}
& \underline{2.26}
& \underline{3.97}
& 0.95
& \underline{3.89}
& \underline{5.42}
& \underline{7.32}
& \underline{0.88}
& \underline{5.19}
& \underline{8.67}
& \underline{12.20}
& 0.67
& 12.19 \\

$\mathcal{L}_{\mathrm{base}}
+\mathcal{L}_{\mathrm{PV}}
+\mathcal{L}_{\mathrm{ord}}$
& \textbf{3.97}
& \textbf{9.29}
& \textbf{0.89}
& \textbf{9.15}
& \textbf{2.14}
& \textbf{3.64}
& \textbf{0.98}
& \textbf{3.42}
& \textbf{4.70}
& \textbf{6.40}
& \textbf{0.91}
& \textbf{4.56}
& \textbf{8.61}
& \textbf{11.95}
& \textbf{0.71}
& \textbf{11.61} \\

\bottomrule
\end{tabular}
\endgroup
\caption{Comparison of numerical supervision objectives under the
MSDG protocol. Each dataset serves once as the unseen target domain,
with the remaining three datasets used as source domains. All variants
retain $\mathcal{L}_{\mathrm{LM}}$ and
$\mathcal{L}_{\mathrm{aux}}$ and differ only in their additional
numerical supervision. Loss weights are omitted from the objective
labels for brevity. The best and second-best results are shown in bold
and underlined, respectively.}
\label{tab:supp_numerical_supervision}
\end{table*}

As shown in Table~\ref{tab:supp_numerical_supervision}, Number Token
Loss alone does not provide consistent cross-domain improvements.
Relative to $\mathcal{L}_{\mathrm{base}}$, it degrades all four metrics
on PURE and UBFC-rPPG, while yielding modest improvements on BUAA and
on RMSE, correlation, and SD for VIPL-HR. This domain-dependent
behavior suggests that position-independent token-distance supervision
does not reliably preserve numerical relationships under different
source-to-target distribution shifts.

Adding Cross-Sample Ordinal Supervision substantially improves the
Number Token Loss variant on PURE and UBFC-rPPG. It also reduces MAE,
RMSE, and SD on BUAA, and reduces MAE and RMSE on VIPL-HR. However,
the resulting objective remains worse than
$\mathcal{L}_{\mathrm{base}}$ on all UBFC-rPPG metrics and does not
consistently improve correlation or SD on the remaining targets.
Explicit cross-sample ordering is therefore beneficial, but cannot
fully compensate for an insufficient within-sample representation of
numerical error.

Place-Value-Aware Numerical Supervision provides substantially more
stable cross-domain improvements. Compared with
$\mathcal{L}_{\mathrm{base}}$, it reduces MAE and RMSE on all four
target domains, improves or maintains correlation on PURE, BUAA, and
VIPL-HR, and reduces SD on PURE, UBFC-rPPG, and BUAA. Correlation
on UBFC-rPPG decreases slightly from 0.96 to 0.95, while SD on
VIPL-HR increases marginally. These results support weighting digit
errors according to their numerical significance rather than treating
all digit positions uniformly.

Combining $\mathcal{L}_{\mathrm{PV}}$ with
$\mathcal{L}_{\mathrm{ord}}$ further improves 15 of the 16
dataset--metric combinations and matches the remaining one. In
particular, MAE decreases from 4.33 to 3.97 on PURE, from 2.26 to 2.14
on UBFC-rPPG, from 5.42 to 4.70 on BUAA, and from 8.67 to 8.61 on
VIPL-HR. The complete Structured Numerical Alignment achieves the
best or tied-best result in all 16 dataset--metric combinations. These
results support the complementarity of the two objectives:
place-value-aware supervision improves within-sample numerical
fidelity, whereas ordinal supervision helps preserve cross-sample HR
ordering.

\subsection{Effect of PhysToken Sequence Length}
\label{sec:supp_phystoken_number}

We study the sensitivity of VITAL to the PhysToken sequence length
under the MSDG protocol by varying the number of PhysTokens
$N_p\in\{5,10,20,40,80,159\}$. Each of PURE, UBFC-rPPG, BUAA, and
VIPL-HR serves once as the unseen target domain, with the remaining
three datasets used as source domains.

The physiological encoder produces a 159-step feature sequence from
the 159 frame differences in each input clip. For $N_p<159$, adaptive
average pooling is applied along the temporal dimension to reduce the
sequence to $N_p$ features. The physiological projector then maps each
pooled feature independently into the language-model hidden space,
producing $N_p$ PhysTokens.

Each configuration is trained independently using the same
physiological encoder, Progressive PhysToken Alignment schedule,
Structured Numerical Alignment objectives, source data, optimization
settings, and total number of training steps. The selected PhysToken
length is used consistently during both training and inference. All
experiments use the Qwen3-VL-4B backbone, and the results are averaged
over five runs with matched random seeds.

\begin{table*}[t]
\centering
\begingroup
\small
\setlength{\tabcolsep}{2pt}
\begin{tabular}{@{}c*{16}{c}@{}}
\toprule
\multirow{2}{*}{\shortstack{Number of\\PhysTokens}}
& \multicolumn{4}{c}{PURE}
& \multicolumn{4}{c}{UBFC-rPPG}
& \multicolumn{4}{c}{BUAA}
& \multicolumn{4}{c}{VIPL-HR} \\
\cmidrule(lr){2-5}
\cmidrule(lr){6-9}
\cmidrule(lr){10-13}
\cmidrule(lr){14-17}
& MAE$\downarrow$
& RMSE$\downarrow$
& $r\uparrow$
& SD$\downarrow$
& MAE$\downarrow$
& RMSE$\downarrow$
& $r\uparrow$
& SD$\downarrow$
& MAE$\downarrow$
& RMSE$\downarrow$
& $r\uparrow$
& SD$\downarrow$
& MAE$\downarrow$
& RMSE$\downarrow$
& $r\uparrow$
& SD$\downarrow$ \\
\midrule

5
& 8.83
& 14.09
& 0.83
& 13.92
& 6.95
& 8.87
& 0.94
& 7.66
& 7.43
& 9.65
& 0.82
& 7.83
& 12.85
& 17.35
& 0.53
& 14.87 \\

10
& \underline{4.27}
& 9.86
& 0.87
& 9.70
& 3.18
& 3.90
& 0.96
& 3.73
& 5.04
& 6.83
& 0.86
& 6.41
& 9.60
& 12.46
& \underline{0.68}
& 12.29 \\

20
& \textbf{3.97}
& \textbf{9.29}
& \textbf{0.89}
& \textbf{9.15}
& \textbf{2.14}
& \textbf{3.64}
& \textbf{0.98}
& \textbf{3.42}
& \textbf{4.70}
& \textbf{6.40}
& \textbf{0.91}
& \textbf{4.56}
& \textbf{8.61}
& \textbf{11.95}
& \textbf{0.71}
& \textbf{11.61} \\

40
& 4.52
& 10.03
& \underline{0.88}
& 9.91
& \underline{2.33}
& \underline{3.87}
& \underline{0.97}
& \underline{3.53}
& \underline{4.86}
& \underline{6.69}
& \underline{0.87}
& \underline{5.19}
& \underline{9.03}
& \underline{12.23}
& 0.65
& \underline{12.05} \\

80
& 4.38
& \underline{9.80}
& 0.87
& \underline{9.54}
& 2.59
& 3.91
& 0.96
& 3.62
& 5.22
& 6.90
& 0.85
& 5.35
& 9.45
& 12.55
& 0.64
& 12.34 \\

159
& 5.04
& 10.32
& 0.85
& 10.29
& 3.28
& 4.12
& 0.95
& 4.10
& 5.47
& 7.39
& 0.83
& 6.60
& 10.00
& 13.58
& 0.61
& 13.46 \\

\bottomrule
\end{tabular}
\endgroup
\caption{Effect of PhysToken sequence length under the MSDG protocol.
Each dataset serves once as the unseen target domain, with the
remaining three datasets used as source domains. The best and
second-best results are shown in bold and underlined, respectively.}
\label{tab:supp_phystoken_number}
\end{table*}

As shown in Table~\ref{tab:supp_phystoken_number}, PhysToken sequence
length exhibits a consistent non-monotonic effect across the four
unseen target domains. Increasing the sequence length from 5 to 10
improves all 16 dataset--metric combinations, indicating that an
overly compressed sequence discards temporal information required for
physiological estimation. Increasing the length from 10 to 20 further
improves all 16 combinations, including a decrease in VIPL-HR RMSE
from 12.46 to 11.95.

The 20-token configuration achieves the best result in all 16
dataset--metric combinations. Increasing the sequence length beyond 20 provides no
additional benefit: the 40-, 80-, and 159-token configurations all
underperform the 20-token model across every metric and target domain.
In particular, retaining the complete 159-step sequence consistently
produces higher errors and lower correlations than temporal
compression to 20 PhysTokens.

These results suggest a trade-off between temporal information
preservation and alignment complexity. Very short sequences may remove
physiologically informative temporal dynamics, whereas excessively
long sequences may retain redundant local variations and increase the
number of physiological tokens that must be aligned with the
language-model space. Under the current architecture and training
protocol, 20 PhysTokens provide the most effective balance and are
therefore used in all other experiments.

\subsection{Tokenizer Alignment and Valid-Pair Coverage}
\label{sec:supp_tokenizer_coverage}

Structured Numerical Alignment applies additional supervision only
to numerical positions and sample pairs with valid character-to-token
correspondence. We therefore quantify the effective training coverage
of Place-Value-Aware Numerical Supervision and Cross-Sample Ordinal
Supervision using the Qwen3-VL tokenizer adopted in our main
experiments.

For Place-Value-Aware Numerical Supervision, we use the same HR
formatting and character-to-token alignment procedure as during
training. A numerical position is considered valid only when its digit
character can be unambiguously associated with the corresponding
language-model output token. The valid digit-position ratio is defined
as

\begin{equation}
R_{\mathrm{digit}}
=
\frac{
\sum_i \sum_{t\in\Omega_i} r_{i,t}
}{
\sum_i |\Omega_i|
},
\label{eq:supp_valid_digit_ratio}
\end{equation}

where $\Omega_i$ denotes the digit positions in the formatted HR
string of sample $i$, and $r_{i,t}$ indicates whether position $t$
has a valid digit-token correspondence. We additionally report the
proportion of samples for which all digit positions are valid.

For Cross-Sample Ordinal Supervision, initial candidate pairs are
first constructed using the predefined rank offsets. Pairs whose
ground-truth HR difference does not exceed the noise threshold are
then removed. The remaining pairs are further filtered by requiring
the formatted HR strings to have the same number of integer digits
and compatible token-span structures at the first differing digit.
Accordingly, the filtering process is

\begin{equation}
\begin{aligned}
\text{Initial Pairs}
&\rightarrow \text{HR-Gap Valid}
\rightarrow \text{Same Length} \\
&\rightarrow \text{Span Compatible}
\rightarrow \text{Final Valid Pairs}.
\end{aligned}
\label{eq:supp_pair_filtering}
\end{equation}

HR-Gap Retention is computed relative to all offset-based initial
candidate pairs. Same-Length Retention is computed relative to
HR-gap-valid pairs, while Span-Compatible Retention is computed
relative to same-length pairs. Final Valid Pairs is computed relative
to all offset-based initial candidate pairs.

All statistics are computed over the source-domain training data of
the four MSDG settings. Pair statistics are averaged over five
dataloader seeds using the same batch size, rank offsets, and
HR-difference threshold as during training. This analysis does not
require model forward or backward computation.

\begin{table*}[t]
\centering
\begingroup
\small
\setlength{\tabcolsep}{3.5pt}
\begin{tabular}{@{}l*{7}{c}@{}}
\toprule
\shortstack[l]{Target\\Domain}
& \shortstack{Valid Digit\\Positions}
& \shortstack{Fully Aligned\\Samples}
& \shortstack{HR-Gap\\Retention}
& \shortstack{Same-Length\\Retention}
& \shortstack{Span-Compatible\\Retention}
& \shortstack{Final Valid\\Pairs}
& \shortstack{Valid Pairs\\per Batch} \\
\midrule

PURE
& 100.00\%
& 100.00\%
& 87.46\%
& 62.14\%
& 100.00\%
& 54.35\%
& 2.72 \\

UBFC-rPPG
& 100.00\%
& 100.00\%
& 87.11\%
& 82.13\%
& 100.00\%
& 71.54\%
& 3.58 \\

BUAA
& 100.00\%
& 100.00\%
& 90.33\%
& 61.75\%
& 100.00\%
& 55.78\%
& 2.79 \\

VIPL-HR
& 100.00\%
& 100.00\%
& 89.98\%
& 66.55\%
& 100.00\%
& 59.88\%
& 2.99 \\

Micro-average
& 100.00\%
& 100.00\%
& 88.38\%
& 68.58\%
& 100.00\%
& 60.61\%
& 3.03 \\

\bottomrule
\end{tabular}
\endgroup
\caption{Coverage analysis of Structured Numerical Alignment using the
Qwen3-VL tokenizer. Valid Digit Positions is computed over all
formatted digit positions. HR-Gap Retention is computed relative to
all offset-based initial candidate pairs, Same-Length Retention
relative to HR-gap-valid pairs, and Span-Compatible Retention relative
to same-length pairs. Final Valid Pairs is computed relative to all
offset-based initial candidate pairs. Pair statistics are averaged
over five dataloader seeds.}
\label{tab:supp_tokenizer_coverage}
\end{table*}

As shown in Table~\ref{tab:supp_tokenizer_coverage}, all formatted
digit positions have valid token correspondence, and every HR sample
is fully aligned under the Qwen3-VL tokenizer. For Cross-Sample
Ordinal Supervision, 88.38\% of the offset-based initial candidate
pairs satisfy the HR-difference requirement. Among these pairs,
68.58\% have the same number of integer digits, and all retained
same-integer-length pairs have compatible token-span structures at
their first differing digit.

After the complete filtering process, 60.61\% of the initially
constructed candidate pairs remain valid. The final valid-pair
coverage ranges from 54.35\% to 71.54\% across the four MSDG settings,
with an average of 3.03 valid ordinal pairs per mini-batch. These
results indicate that the proposed numerical objectives provide
substantial training coverage across all MSDG settings rather than
relying on a small subset of exceptionally aligned numerical positions
or sample pairs.

\subsection{Sensitivity Analysis of Pairwise Numerical Collisions}
\label{sec:supp_collision_sensitivity}

To further quantify the repeated-value behavior observed in
autoregressive HR generation, we analyze pairwise numerical
collisions under different ground-truth and prediction-distance
thresholds on PURE. The continuous regression probe and
autoregressive variants use the same projector-pretrained
physiological representation.

For a minimum ground-truth HR separation $\gamma$ and prediction
distance threshold $\delta$, the pairwise collision rate is defined as

\begin{equation}
C(\gamma,\delta)
=
\frac{
\sum_{i<j}
\mathbb{I}\left(|h_i-h_j|\geq\gamma\right)
\mathbb{I}\left(|\hat{h}_i-\hat{h}_j|\leq\delta\right)
}{
\sum_{i<j}
\mathbb{I}\left(|h_i-h_j|\geq\gamma\right)
}.
\label{eq:supp_collision_rate}
\end{equation}

Before computing collisions, predictions from all methods are
converted to the same one-decimal precision used by autoregressive
generation. All evaluated outputs are valid and numerically parsable. This metric measures the proportion of sample pairs with clearly
different ground-truth HR values that are mapped to identical or
nearly identical predictions.

\begin{table}[t]
\centering
\begingroup
\small
\setlength{\tabcolsep}{3.5pt}
\begin{tabular}{@{}lcccc@{}}
\toprule
Method
& $\delta=0$
& $\delta=0.5$
& $\delta=1$
& $\delta=2$ \\
\midrule
\shortstack[l]{Continuous\\Regression Probe}
& \textbf{0.72\%}
& \textbf{2.17\%}
& \textbf{3.62\%}
& \textbf{6.64\%} \\

\shortstack[l]{Autoregressive\\Generation}
& 3.16\%
& 5.96\%
& 6.55\%
& 10.01\% \\

\shortstack[l]{Autoregressive\\Generation + SNA}
& \underline{1.85\%}
& \underline{3.72\%}
& \underline{4.73\%}
& \underline{8.49\%} \\
\bottomrule
\end{tabular}
\endgroup
\caption{Pairwise collision rates under different prediction-distance
thresholds. The minimum ground-truth HR separation is fixed at
10 BPM. Lower values indicate better cross-sample numerical
discriminability. Best results are bolded, and second-best results
are underlined.}
\label{tab:supp_prediction_threshold}
\end{table}

As shown in Table~\ref{tab:supp_prediction_threshold}, standard
autoregressive generation produces consistently higher collision
rates than the continuous regression probe across all
prediction-distance thresholds. Under exact numerical equality, its
collision rate increases from 0.72\% to 3.16\%, indicating a
substantially stronger tendency to map physiologically distinct videos
to the same HR value. The difference remains evident when the
allowable prediction distance is increased to 0.5, 1, or 2 BPM,
showing that the observed numerical concentration is not determined
by a particular collision threshold.

Structured Numerical Alignment consistently reduces the collision
rate of autoregressive generation across all settings. For example,
at $\delta=1$ BPM, SNA reduces the collision rate from 6.55\% to
4.73\%, while exact-value collisions decrease from 3.16\% to 1.85\%.
Although the SNA variant does not fully match the continuous
regression probe, its stable improvement indicates that explicitly
modeling numerical significance and cross-sample ordering mitigates
the repeated-value behavior of autoregressive generation.

\begin{table}[t]
\centering
\begingroup
\small
\setlength{\tabcolsep}{3.5pt}
\begin{tabular}{@{}lcccc@{}}
\toprule
Method
& $\gamma=5$
& $\gamma=10$
& $\gamma=15$
& $\gamma=20$ \\
\midrule
\shortstack[l]{Continuous\\Regression Probe}
& \textbf{6.01\%}
& \textbf{3.62\%}
& \textbf{2.20\%}
& \textbf{1.19\%} \\

\shortstack[l]{Autoregressive\\Generation}
& 7.12\%
& 6.55\%
& 5.77\%
& 4.02\% \\

\shortstack[l]{Autoregressive\\Generation + SNA}
& \underline{6.32\%}
& \underline{4.73\%}
& \underline{4.31\%}
& \underline{3.67\%} \\
\bottomrule
\end{tabular}
\endgroup
\caption{Pairwise collision rates under different minimum
ground-truth HR separations. The prediction-distance threshold is
fixed at 1 BPM. Best results are bolded, and second-best results
are underlined.}
\label{tab:supp_gt_threshold}
\end{table}

Table~\ref{tab:supp_gt_threshold} further shows that the relative
disparity between continuous regression and standard autoregressive
generation becomes more pronounced for pairs with larger
ground-truth HR differences. When $\gamma$ increases from 5 to
20 BPM, the collision rate of the continuous regression probe
decreases from 6.01\% to 1.19\%, whereas standard autoregressive
generation retains a collision rate of 4.02\%. Thus, the generative
interface remains more likely to produce similar predictions even
for physiologically distinct inputs.

SNA consistently lowers the collision rate relative to standard
autoregressive generation at every ground-truth separation. Its
improvement is particularly evident at $\gamma=10$ BPM, where the
collision rate decreases from 6.55\% to 4.73\%. Together with the
prediction-threshold analysis, these results show that Structured
Numerical Alignment reduces numerical collisions across different
definitions of pairwise similarity. This supports its effectiveness
in mitigating, although not completely eliminating, the
source-prior attraction and repeated-value banding observed under
distribution shift.

\section{Additional Controls and Cross-Domain Analyses}
\label{sec:supp_additional_controls}

\subsection{Training Stages and Supervision}
\label{sec:supp_training_overview}

Table~\ref{tab:supp_training_overview} summarizes the inputs,
optimized modules, and objectives of the three training stages.
For each target domain, the physiological encoder is pretrained using
only the corresponding source domains. Target-domain data are not used
to train the encoder or adapt VITAL. We keep the physiological encoder
frozen during adaptation to support efficient adaptation and reuse.
An additional test that jointly fine-tunes the physiological encoder,
projector, and VLM gives comparable HR performance, supporting this
choice of adaptation schedule.
The physiological projector and the language-model LoRA parameters
provide the trainable interface between physiological features and
text generation.

\begin{table*}[t]
\centering
\small
\begin{tabular}{p{0.55in}p{1.20in}p{1.24in}p{1.53in}p{1.34in}}
\toprule
Stage & Inputs & Optimized modules & Frozen modules & Active objectives \\
\midrule
Warm-up
& Physiological features
& Physiological projector and auxiliary HR head
& Physiological encoder
& $\mathcal{L}_{\mathrm{aux}}$ \\
Stage I
& PhysTokens and HR instruction
& Physiological projector and LM LoRA
& Physiological encoder and auxiliary HR head
& $\mathcal{L}_{\mathrm{HR}}$ \\
Stage II
& Semantic tokens, PhysTokens, and instruction
& Physiological projector and LM LoRA
& Both visual encoders, native semantic projector, and auxiliary HR head
& $\mathcal{L}_{\mathrm{HR}}$ for HR; $\mathcal{L}_{\mathrm{LM}}$ for general video QA \\
\bottomrule
\end{tabular}
\caption{Training stages. The native semantic pathway is not used in
warm-up or Stage I. Stage II mixes rPPG and ActivityNet-QA samples at a
$4{:}1$ ratio. The numbers of epochs are 60, 20, and 10, respectively.
Here $\mathcal{L}_{\mathrm{HR}}=\mathcal{L}_{\mathrm{LM}}
+\lambda_{\mathrm{PV}}\mathcal{L}_{\mathrm{PV}}
+\lambda_{\mathrm{ord}}\mathcal{L}_{\mathrm{ord}}
+\lambda_{\mathrm{aux}}\mathcal{L}_{\mathrm{aux}}$.
The HR-objective weights are $\lambda_{\mathrm{PV}}=0.5$,
$\lambda_{\mathrm{ord}}=0.1$, and $\lambda_{\mathrm{aux}}=0.1$.}
\label{tab:supp_training_overview}
\end{table*}

The auxiliary loss compares the ground-truth HR with the scalar HR
regressed from PhysTokens. Its purpose is to preserve continuously
decodable HR information during projection into the language-model
space. Freezing the auxiliary head fixes its parameters, but still
allows gradients through that head to update the physiological
projector. This loss therefore supervises the projected representation
without updating the frozen physiological encoder during adaptation.

The numerical and language-modeling objectives have complementary
roles. $\mathcal{L}_{\mathrm{PV}}$ measures numerical error using
probabilities conditioned on the digit-token subset. It is not a
full-vocabulary numerical likelihood. $\mathcal{L}_{\mathrm{LM}}$
supervises the response tokens over the complete vocabulary, including
the response format and non-digit tokens.
$\mathcal{L}_{\mathrm{ord}}$ additionally constrains the relative digit
preferences of distinct videos at their first differing HR digit.

\subsection{Alternative Physiological Encoder}
\label{sec:supp_alternative_encoder}

To evaluate whether VITAL's benefit is restricted to FactorizePhys,
we replace the physiological encoder with
PhysNet~\cite{yu2019remote} and use the four-target MSDG protocol
of the main paper. The other three datasets provide the source
domains for each held-out target.
Table~\ref{tab:supp_physnet_replacement} compares VITAL with PhysNet
against the standalone PhysNet baseline.

\begin{table}[t]
\centering
\small
\setlength{\tabcolsep}{3.5pt}
\begin{tabular}{lrrrr}
\toprule
Method & PURE & UBFC & BUAA & VIPL \\
\midrule
PhysNet & 10.96 & 8.09 & 6.04 & 14.81 \\
VITAL + PhysNet & \textbf{8.27} & \textbf{7.90} & \textbf{5.93} & \textbf{12.84} \\
\bottomrule
\end{tabular}
\caption{Replacing the physiological encoder with PhysNet.
Values are MAE in BPM; lower is better.
UBFC and VIPL denote UBFC-rPPG and VIPL-HR.}
\label{tab:supp_physnet_replacement}
\end{table}

VITAL improves over PhysNet alone on all four targets.
These results extend the evidence beyond the FactorizePhys
configuration, although the size of the improvement depends on the
target domain and the encoder.

\subsection{Comparison with NTIL}
\label{sec:supp_ntil}

In addition to the NTL and NTL-plus-ordinal comparisons in
Table~\ref{tab:supp_numerical_supervision}, we evaluate NTIL
for decimal HR prediction under the same MSDG target splits.
NTIL~\cite{fei2025advancing} incorporates numerical structure
within a generated sequence. VITAL's cross-sample ordinal objective
adds supervision on the HR ordering of different videos at the first
differing digit, complementing within-sample numerical supervision.

\begin{table}[t]
\centering
\small
\setlength{\tabcolsep}{4pt}
\begin{tabular}{lrrrr}
\toprule
Numerical objective & PURE & UBFC & BUAA & VIPL \\
\midrule
NTIL & 5.29 & 3.62 & 6.75 & 8.73 \\
SNA (VITAL) & \textbf{3.97} & \textbf{2.14} & \textbf{4.70} & \textbf{8.61} \\
\bottomrule
\end{tabular}
\caption{NTIL comparison for HR prediction. Values are MAE in BPM
under the four-target MSDG protocol.}
\label{tab:supp_ntil_comparison}
\end{table}

Table~\ref{tab:supp_ntil_comparison} reports lower MAE with SNA
on each target. Together with the NTL-plus-ordinal results, this
comparison supports the value of combining within-sample numerical
alignment with cross-video ordering.

\subsection{Decoding the Final PhysTokens}
\label{sec:supp_final_auxiliary}

After the complete VITAL adaptation, we evaluate the original auxiliary
HR head on the final PhysTokens and compare its predictions with the
autoregressive language output. The auxiliary head is the head learned
during warm-up and kept frozen during the subsequent adaptation stages;
it is not refitted after the projector has completed adaptation.

\begin{table}[t]
\centering
\small
\setlength{\tabcolsep}{3.5pt}
\begin{tabular}{lrrrr}
\toprule
Readout & PURE & UBFC & BUAA & VIPL \\
\midrule
Frozen auxiliary head & 9.47 & 5.96 & 7.16 & 10.54 \\
Language output & \textbf{3.97} & \textbf{2.14} & \textbf{4.70} & \textbf{8.61} \\
\bottomrule
\end{tabular}
\caption{HR readout after full VITAL training. Values are MAE in BPM.
The auxiliary head is frozen after warm-up, whereas the physiological
projector and language-model LoRA parameters continue to adapt.}
\label{tab:supp_final_auxiliary}
\end{table}

The language output has lower MAE on all four targets in this
configuration. The result characterizes the final VITAL readouts
under their actual training schedule; the auxiliary objective serves
as a representation constraint during adaptation rather than the
inference-time prediction interface.

\subsection{Semantic Pathway and QA Mixing}
\label{sec:supp_semantic_qa}

We examine the contributions of the semantic pathway and general
video QA mixing using MVBench. The physiological-only variant is
trained without the semantic pathway, rather than removing that
pathway only at inference. A separate variant retains the VITAL
architecture but omits general QA mixing during adaptation.
All variants in Table~\ref{tab:supp_semantic_qa} use Qwen3-VL-4B.

\begin{table}[t]
\centering
\small
\begin{tabular}{lr}
\toprule
Model & MVBench accuracy (\%) \\
\midrule
Native Qwen3-VL-4B & 64.05 \\
VITAL without semantic pathway & 37.29 \\
VITAL without QA mixing & 56.57 \\
Full VITAL & 60.53 \\
\bottomrule
\end{tabular}
\caption{General video understanding under semantic-pathway and
QA-mixing ablations. Full VITAL uses $4{:}1$ rPPG-to-QA mixing.}
\label{tab:supp_semantic_qa}
\end{table}

The complete model exceeds the physiological-only variant by
23.24 percentage points. QA mixing improves MVBench accuracy
from 56.57\% to 60.53\%, a gain of 3.96 percentage points.
The final accuracy remains 3.52 percentage points below the native
model, indicating a remaining adaptation cost despite the benefit
of retaining the semantic pathway and including general video QA.

\subsection{Collision Rates after Full Adaptation}
\label{sec:supp_full_collision}

The diagnostic probes in Tables~\ref{tab:supp_prediction_threshold}
and~\ref{tab:supp_gt_threshold} isolate
output-interface behavior from a shared physiological representation.
Here we additionally compare fully adapted VITAL models with and
without SNA under matched training settings.
We use the collision rate in Eq.~\ref{eq:supp_collision_rate}, fix
the prediction-distance threshold to $\delta=1$ BPM, and vary the
minimum ground-truth separation over $\gamma\in\{5,10,15,20\}$ BPM.
The 1-BPM tolerance identifies nearly identical numerical predictions;
Table~\ref{tab:supp_prediction_threshold} separately examines the
sensitivity of the diagnostic result to alternative tolerances.

\begin{table*}[t]
\centering
\small
\begin{tabular}{llrrrr}
\toprule
Training and target setting & Variant & $\gamma=5$ & $\gamma=10$ & $\gamma=15$ & $\gamma=20$ \\
\midrule
UBFC-rPPG $\rightarrow$ PURE & Without SNA & 4.768 & 4.476 & 4.445 & 3.367 \\
& Full VITAL & \textbf{3.481} & \textbf{2.754} & \textbf{2.304} & \textbf{1.460} \\
\midrule
MSDG $\rightarrow$ PURE & Without SNA & 1.486 & 1.190 & 1.175 & 1.280 \\
& Full VITAL & \textbf{0.720} & \textbf{0.526} & \textbf{0.473} & \textbf{0.416} \\
\midrule
MSDG $\rightarrow$ BUAA & Without SNA & 3.752 & 2.468 & 1.196 & 0.255 \\
& Full VITAL & \textbf{1.657} & \textbf{1.053} & \textbf{0.444} & \textbf{0.112} \\
\bottomrule
\end{tabular}
\caption{Collision rates (\%) of fully adapted VITAL, with prediction
differences at most 1 BPM. Each column conditions on ground-truth
differences of at least $\gamma$ BPM. MSDG uses the other three
datasets among PURE, UBFC-rPPG, BUAA, and VIPL-HR as sources.
The better result within each setting is bolded.}
\label{tab:supp_full_collision}
\end{table*}

SNA lowers the collision rate at every evaluated separation in
Table~\ref{tab:supp_full_collision}. At $\gamma=20$ BPM on
UBFC-rPPG $\rightarrow$ PURE, the rate decreases from 3.367\%
to 1.460\%, a relative reduction of 56.6\%.
The corresponding MSDG rates decrease from 1.280\% to 0.416\%
on PURE and from 0.255\% to 0.112\% on BUAA.
These comparisons quantify reduced concentration of predictions
for samples with distinct true HR values after full adaptation.
They do not independently separate HR-prior changes from other
changes between datasets.

\subsection{Source and Target HR Statistics}
\label{sec:supp_hr_distributions}

\begin{figure}[t]
\centering
\includegraphics[width=\columnwidth]{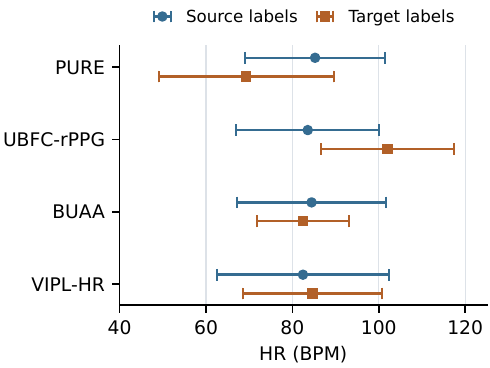}
\caption{Source and target HR label means with one-SD horizontal
intervals for each MSDG setting. The intervals describe HR dispersion,
not confidence intervals for the means.}
\label{fig:supp_hr_summary}
\end{figure}

Table~\ref{tab:supp_hr_source_target} characterizes the source and
target HR labels for the four MSDG settings. Source summaries are
reconstructed from the saved training configurations and available
data manifests, with each available clip counted once after filtering.
These are clip-level summaries, rather than subject-balanced
statistics or a reconstruction of stochastic sampler frequencies.
Here, SD describes the dispersion of HR values; it is distinct from
the standard deviation of prediction errors used in the performance
tables. Figure~\ref{fig:supp_hr_summary} visualizes the label means
and SDs without assuming a distributional shape.

\begin{table*}[t]
\centering
\small
\begin{tabular}{llrrrr}
\toprule
Held-out target & Label set & Clips & Median & IQR & Mean $\pm$ SD \\
\midrule
PURE & Sources: B+V+U & 48,682 & 82.60 & 22.40 & $85.25\pm16.16$ \\
& Target & 3,111 & 67.90 & 18.70 & $69.32\pm20.28$ \\
\midrule
UBFC-rPPG & Sources: P+B+V & 49,703 & 81.20 & 21.90 & $83.55\pm16.51$ \\
& Target & 2,090 & 103.30 & 21.10 & $101.99\pm15.47$ \\
\midrule
BUAA & Sources: P+V+U & 48,036 & 81.80 & 24.20 & $84.44\pm17.25$ \\
& Target & 3,757 & 83.10 & 9.70 & $82.42\pm10.67$ \\
\midrule
VIPL-HR & Sources: P+B+U & 8,958 & 81.20 & 25.50 & $82.44\pm19.89$ \\
& Target & 42,835 & 81.80 & 22.30 & $84.68\pm16.13$ \\
\bottomrule
\end{tabular}
\caption{Source and target HR label summaries under MSDG.
P, U, B, and V denote PURE, UBFC-rPPG, BUAA, and VIPL-HR,
respectively. All HR statistics are in BPM. IQR is the
interquartile range.}
\label{tab:supp_hr_source_target}
\end{table*}

The source-to-target mean changes differ by target: PURE has
85.25 versus 69.32 BPM, whereas UBFC-rPPG has 83.55 versus
101.99 BPM. BUAA and VIPL-HR exhibit smaller mean shifts,
although their dispersions and quartile statistics also differ.
Thus, a single shift pattern does not characterize all settings.

\begin{table*}[t]
\centering
\small
\begin{tabular}{llrrrr}
\toprule
Target & Output & Clips & Median & IQR & Mean $\pm$ SD \\
\midrule
PURE & Without SNA & 3,111 & 69.85 & 17.93 & $71.95\pm17.80$ \\
& Full VITAL & 3,111 & 68.71 & 18.63 & $70.92\pm18.40$ \\
& Ground truth & 3,111 & 67.90 & 18.70 & $69.32\pm20.28$ \\
\midrule
UBFC-rPPG & Without SNA & 2,090 & 100.87 & 22.66 & $100.83\pm15.51$ \\
& Full VITAL & 2,090 & 102.41 & 20.19 & $100.76\pm14.81$ \\
& Ground truth & 2,090 & 103.30 & 21.10 & $101.99\pm15.47$ \\
\midrule
BUAA & Without SNA & 3,757 & 75.68 & 9.75 & $76.22\pm9.44$ \\
& Full VITAL & 3,757 & 78.33 & 8.75 & $77.93\pm10.29$ \\
& Ground truth & 3,757 & 83.10 & 9.70 & $82.42\pm10.67$ \\
\midrule
VIPL-HR & Without SNA & 42,835 & 83.10 & 17.90 & $86.49\pm14.57$ \\
& Full VITAL & 42,835 & 85.00 & 17.93 & $86.33\pm14.56$ \\
& Ground truth & 42,835 & 81.80 & 22.30 & $84.68\pm16.13$ \\
\bottomrule
\end{tabular}
\caption{Prediction and ground-truth HR summaries on each MSDG target.
The same target clips are used within each block. Units are BPM.
SD here refers to the distribution of the displayed HR values,
not the distribution of prediction errors.}
\label{tab:supp_hr_predictions}
\end{table*}

Table~\ref{tab:supp_hr_predictions} reports prediction summaries
on the target clips. As a concrete example, on PURE the mean signed
error decreases from $71.95-69.32=+2.63$ BPM without SNA
to $70.92-69.32=+1.60$ BPM with SNA.
This observation is consistent with reduced attraction toward the
higher source HR mean in this setting. The cross-dataset comparisons
remain observational and do not establish the HR prior as the sole
cause of prediction bias.

\FloatBarrier
\subsection{Illumination-Stratified Analysis on BUAA}
\label{sec:supp_buaa_illumination}

We compare the standalone FactorizePhys PPG-to-HR pipeline with full
VITAL across the six evaluated BUAA illumination conditions.
Table~\ref{tab:supp_buaa_illumination} reports the stratified MAE.
FactorizePhys predicts a physiological waveform from which HR is
computed, whereas VITAL generates the HR through its language output.

\begin{table}[!htbp]
\centering
\small
\begin{tabular}{rrr}
\toprule
Illumination (lux) & FactorizePhys & VITAL \\
\midrule
10.0 & \textbf{4.876} & 5.368 \\
15.8 & \textbf{3.532} & 4.491 \\
25.1 & \textbf{2.797} & 5.114 \\
39.8 & \textbf{2.297} & 4.269 \\
63.1 & \textbf{2.784} & 4.782 \\
100.0 & \textbf{2.169} & 4.234 \\
\midrule
Overall & \textbf{3.060} & 4.702 \\
\bottomrule
\end{tabular}
\caption{MAE (BPM) by illumination on BUAA.
FactorizePhys uses waveform prediction followed by HR estimation;
VITAL uses its autoregressive language output.
The overall row is the reported aggregate for this stratified analysis.}
\label{tab:supp_buaa_illumination}
\end{table}

VITAL has higher MAE at all six illumination levels.
The gap is smallest at 10 lux (0.492 BPM), compared with
0.959--2.317 BPM at the other evaluated levels.
Both methods have their highest MAE at 10 lux, but the additional
gap between VITAL and FactorizePhys is not amplified there.
The difference is therefore not confined to the lowest illumination.
These observations do not establish that the same gap will occur
under low light in other datasets.

\subsection{Clarification of Evaluation Settings}
\label{sec:supp_evaluation_clarification}

For prediction errors $e_i=\widehat{h}_i-h_i$ within one evaluation
run, RMSE is $\sqrt{\operatorname{mean}(e_i^2)}$ and SD is
$\operatorname{std}(e_i)$. SD measures the spread of prediction
errors within a run, rather than variability across training seeds.

The Period-LLM-style baseline uses Qwen3-VL-4B in both the HR-only
and general-capability experiments. Its main-paper HR-estimation
comparisons use HR-only fine-tuning, while the general-capability
comparison retrains it with $4{:}1$ rPPG-to-QA mixing.
The MVBench and HR results for that mixed variant come from the same
set of training runs. VITAL uses the mixed Stage-II setting described
in Table~\ref{tab:supp_training_overview}.

The main-paper joint semantic and physiological evaluation uses a
single multi-question prompt requesting semantic attributes, recovery
state, and HR together. This setting is distinguished from the
standalone HR question used in the primary HR evaluation.
The auxiliary regression head is used only during training and
in the explicit readout analysis of
Table~\ref{tab:supp_final_auxiliary}; the standard VITAL inference
interface generates HR through the original language head.